\documentclass[]{bytedance_seed}

\usepackage[toc,page,header]{appendix}

\usepackage{amssymb}
\usepackage{amsfonts}
\usepackage{xspace}
\usepackage{minitoc}
\usepackage{graphicx}
\usepackage{xcolor}
\usepackage{wrapfig}
\usepackage{color, colortbl}

\newcommand{\grayt}{\color{black}}
\newcommand{\fst}[1]{\textbf{#1}}

\definecolor{themecolor}{HTML}{C7DAFF}
\newcommand{\graybg}{\rowcolor{themecolor!50}}
\newcommand{\ourmodel}{Stable-DiffCoder\xspace}

\usepackage{amsmath}
\usepackage{mathtools}
\newtheorem{theorem}{Theorem}[section]
\newtheorem{definition}[theorem]{Definition}

\title{Stable-DiffCoder: Pushing the Frontier of Code Diffusion Large Language Model}

\author[1,2]{Chenghao Fan}
\author[2]{Wen Heng}
\author[2]{Bo Li}
\author[1]{Sichen Liu}
\author[2]{Yuxuan Song}
\author[2]{Jing Su}
\author[1]{Xiaoye Qu}
\author[2]{Kai Shen}
\author[1]{Wei Wei}

\affiliation[1]{Huazhong University of Science and Technology}
\affiliation[2]{ByteDance Seed}

\abstract{
Diffusion-based language models (DLLMs) offer non-sequential, block-wise generation and richer data reuse compared to autoregressive (AR) models, but existing code DLLMs still lag behind strong AR baselines under comparable budgets. 
We revisit this setting in a controlled study and introduce \ourmodel, a block diffusion code model that reuses the Seed-Coder architecture, data, and training pipeline. To enable efficient knowledge learning and stable training, we incorporate a block diffusion continual pretraining (CPT) stage enhanced by a tailored warmup and block-wise clipped noise schedule. Under the same data and architecture, \ourmodel overall outperforms its AR counterpart on a broad suite of code benchmarks. Moreover, relying only on the CPT and supervised fine-tuning stages, \ourmodel achieves stronger performance than a wide range of
$\sim$8B ARs and DLLMs, demonstrating that diffusion-based training can improve code modeling
quality beyond AR training alone. 
Moreover, diffusion-based any-order modeling improves structured code modeling for editing and reasoning, and through data augmentation, benefits low-resource coding languages.
}
\date{\today}
\correspondence{Wei Wei at \email{weiw@hust.edu.cn}}

\checkdata[Project Page]{\url{https://github.com/ByteDance-Seed/Stable-DiffCoder}}
\checkdata[Model Page]{\url{https://huggingface.co/collections/ByteDance-Seed/stable-diffcoder}}

\begin{document}
\maketitle
\begin{figure}[h]
\vspace{-15pt}
    \centering
    \includegraphics[
        width=0.98\columnwidth,
        trim=20pt 650pt 20pt 20pt,
        clip
    ]{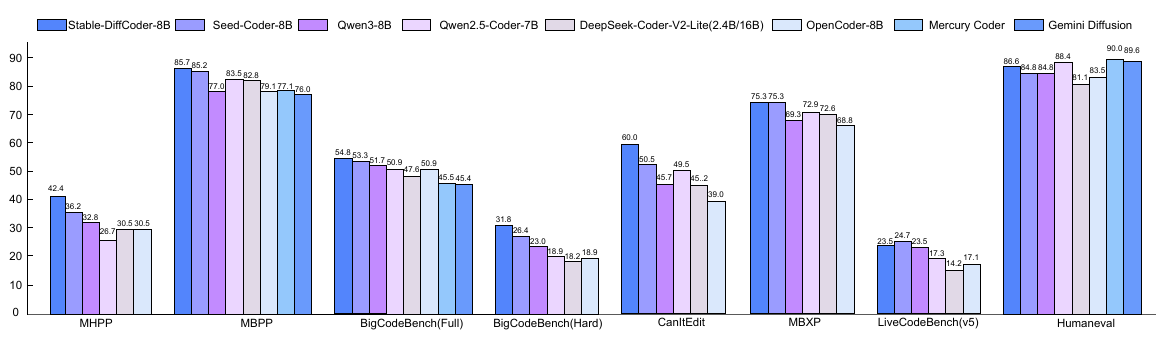}
    \caption{
     Benchmark performance of \ourmodel-8B-Instruct.
    }
    \label{fig:ablation-settings}
\end{figure}
\newpage

\section{Introduction}

Autoregressive (AR) language models have achieved strong results in natural language and code by modeling sequences left‑to‑right~\citep{dubey2024llama,liu2024deepseek,seed2025seed,yang2025qwen3,huang2025opencoder,hui2024qwen2,zhu2024deepseek}, but their strictly sequential decoding underutilizes the inherently non‑autoregressive nature of code, where developers routinely infill missing spans, iterate from a scratch toward the final version, revise earlier segments using later context, and generate independent blocks in parallel~\citep{li2025beyond,Dreamon2025,israel2025planned}. To better align with this paradigm, diffusion‑based language models (DLLMs)~\citep{nie2025largelanguagediffusionmodels,austin2023structureddenoisingdiffusionmodels} provide a complementary view by treating generation as iterative denoising: they corrupt and reconstruct sequences under random masking, enabling non‑sequential, block‑wise decoding and implicitly augmenting each clean example with many arbitrary corruption patterns, which is especially appealing for rare high‑quality or long‑tail code samples~\citep{ni2025diffusion,gao2025makes}.
Recently, mask‑based DLLMs that instantiate this corruption‑and‑denoising view have gained attention for their potential in fast parallel generation and high inference speed ceilings~\citep{song2025seed,wu2025fastdllmtrainingfreeaccelerationdiffusion,liu2025wedlmreconcilingdiffusionlanguage,xu2025lopa,monsefi2025fs,ma2025dinfer}. In this work, we therefore focus on such mask‑based DLLMs and use ``training modes'' to refer to the different corruption or masking patterns sampled for each example. 
Despite this focus on efficiency, their impact on modeling quality remains unclear, leaving open whether diffusion can systematically improve model capability. ln principle, this stochastic training should help models absorb scarce supervision and generalize beyond specific surface forms, yet existing diffusion‑based code models still lag behind strong AR baselines in overall accuracy~\cite{dream2025,song2025seed,kang2025parallelbench}, and prior work often changes data, architecture, and training pipeline simultaneously~\cite{xie2025dream,wu2025fast}, leaving open a central question: \textit{under a fixed data and compute budget, can the additional training modes introduced by diffusion actually improve model capability?} 


In this work, we systematically study how to enable diffusion models to learn new knowledge more efficiently, by conducting controlled experiments on a 2.5B model size.
Building on the observations, we propose \ourmodel, a diffusion language model that reuses the autoregressive Seed‑Coder~\citep{seed2025seed} pipeline and data but achieves stronger performance. Starting from the AR Seed‑Coder checkpoint before annealing, we perform continual pretraining on 1.3T tokens using small‑block diffusion (block size 4), preserving representation plasticity while introducing multi‑token prediction patterns. To make the training process more suitable for DLLM, we design (i) the tailored warmup training process, (ii) a block‑wise clipped noise schedule that guarantees non‑trivial supervision within each block. 
Finally, we carry out a comprehensive evaluation on a broad suite of code benchmarks. Across both base and instruction‑tuned settings, \ourmodel almost uniformly surpasses its AR counterpart Seed‑Coder of the same size, and establishes new state‑of‑the‑art results among ~8B‑scale diffusion code models on many metrics. These results suggest that, when equipped with an appropriate curriculum and training design, diffusion language models can not only match but exceed autoregressive models in code understanding and generation quality, even before factoring in any potential inference‑time speed advantages.

\section{Preliminary}
\subsection{Autoregressive Language Models}
Transformer-based~\citep{vaswani2017attention} AR models have been the dominant paradigm in language modeling, powering many state-of-the-art code and text generation systems~\citep{dubey2024llama,liu2024deepseek,seed2025seed,yang2025qwen3,huang2025opencoder,hui2024qwen2,zhu2024deepseek}.
Let $\mathcal{V}$ be a finite vocabulary and $\mathbf{x}_{1:T}=(x_1,\dots,x_T)\in\mathcal{V}^T$ a token sequence. In an AR language model, it is assumed that each token is conditioned on all preceding tokens, so that the joint distribution is factorized in a left-to-right manner as
\begin{equation}
p_\theta(\mathbf{x}_{1:N}) = p_\theta(\mathbf{x}_1)\prod_{i=2}^N p_\theta(\mathbf{x}_i \mid \mathbf{x}_{<i})
\label{eq:ar-factorization}
\end{equation}
where $x_t$ is the $t$-th token in the sequence. We typically parameterize $p_\theta$ with a Transformer, and train by maximizing token-level cross-entropy. Concretely, given data ${x}\sim p_{\text{data}}$, the training loss is:
\begin{equation}
\mathcal{L}_{\mathrm{AR}}(\theta) = -\mathbb{E}_{\mathbf{x}\sim p_{\mathrm{data}}}\Big[\log p_\theta(\mathbf{x}_1)+\sum_{i=2}^N \log p_\theta(\mathbf{x}_i \mid \mathbf{x}_{<i})\Big]
\label{eq:ar-loss}
\end{equation}
This is the standard next-token prediction objective widely used in mainstream AR models.
Notable AR transformers like GPT-series~\cite{brown2020language,yang2025qwen3,liu2024deepseek} and code-specific LLMs~\citep{seed2025seed,hui2024qwen2} have achieved high accuracy on programming benchmarks, but their token-by-token generation can be slow and lacks a mechanism for global planning or simultaneous reasoning over multiple parts of the sequence. This limitation has motivated exploration of non-AR generation alternatives.

\subsection{Masked Diffusion Language Models}
Masked diffusion language models (DLLMs) have recently emerged as a promising non-AR alternative for text and code generation~\citep{nie2025largelanguagediffusionmodels,zhu2025llada15variancereducedpreference,dream2025,bie2025llada2,cheng2025sdar,inception2025mercury,gemini_diffusion_2025,song2025seed}. Inspired by continuous diffusion models in vision, discrete diffusion LMs optimize a variational evidence lower bound by corrupting observed sequences (e.g., by randomly masking tokens) and training a denoiser to reconstruct the clean data~\citep{austin2023structureddenoisingdiffusionmodels,loudiscrete,meng2022concrete}. 
This framework provides a principled likelihood-based foundation for parallel or partially parallel text generation, which earlier heuristic non-AR methods often lacked~\citep{Dreamon2025,liu2025tidar,kim2025train}.

Instead of factorizing the distribution autoregressively, DLLMs define a sequence of latent variables 
$\mathbf{x}_{1:N}^0,\mathbf{x}_{1:N}^1,..$ $.,\mathbf{x}_{1:N}^T$, 
where $\mathbf{x}_{1:N}^0$ denotes the clean text and $\mathbf{x}_{1:N}^T$ denotes a fully corrupted state. In DLLMs, the corruption is instantiated via a dedicated \textsc{[MASK]} token. The forward noising process progressively replaces tokens with \textsc{[MASK]}, while the reverse denoising process iteratively reconstructs the clean text step by step.
\begin{equation}
p_\theta(\mathbf{x}_{1:N}) = p_\theta(\mathbf{x}_{1:N}^T)\prod_{t=1}^T p_\theta(\mathbf{x}_{1:N}^{t-1}\mid \mathbf{x}_{1:N}^t)
\label{eq:dlm-factorization}
\end{equation}
where $p_\theta(\mathbf{x}_{1:N}^T)$ is typically a uniform prior over noise states, and $p_\theta(\mathbf{x}_{1:T}^{t-1}\mid \mathbf{x}_{1:T}^t)$ is parameterized by a neural network. 

In particular, for DLLMs, prior theoretical analyses \citep{sahoo2024simpleeffectivemaskeddiffusion,ou2025absorbingdiscretediffusionsecretly,shi2024simplified} have shown that this variational denoising objective can be equivalently reduced to a weighted cross-entropy loss, which enables efficient training in practice:
\begin{equation}
\mathcal{L}_{\mathrm{DLLM}}(\theta) = -\mathbb{E}_{\mathbf{x^0}\sim p_{\mathrm{data}},\, t\sim \mathcal{U}(0,1),\, \mathbf{x}^t\sim q(\mathbf{x}^t\mid \mathbf{x}^0)}
\Big[w(t)\sum_{i=1}^N   {1}[\mathbf{x}^t_{i}=\text{MASK}]\,\log p_\theta(\mathbf{x}^0_i\mid \mathbf{x}^t_{1:N} )\Big]
\label{eq:dlm-loss}
\end{equation}
where $q(\mathbf{x}^t \mid \mathbf{x}^0)$ denotes the corruption process, ${1}[\mathbf{x}^t_i=\text{[MASK]}]$ is the indicator function specifying whether the $i$-th token is masked at timestep $t$, and $w(t)$ is a weighting coefficient that depends on the corruption level.
For example, under a linear noise schedule one may set $w(t) = \tfrac{1}{t}$.
This formulation allows DLLMs to leverage a wide range of advanced model architectures, requiring only minor modifications to the original implementation in order to support diffusion language model's training. 
Recently, several works combine AR structure with diffusion to form block diffusion variants of DLLMs~\citep{arriola2025block,bie2025llada2,cheng2025sdar,liu2025sequential,tian2025next,wu2025fastdllmtrainingfreeaccelerationdiffusion,wu2025fast,liu2025tidar}. Instead of corrupting the whole sequence with random masks, these models only diffuse a contiguous block of tokens at each step while keeping the surrounding context clean, so that the denoiser learns to generate a span conditioned on a mostly uncorrupted prefix. 

Beyond offering a higher ceiling for parallel decoding speed~\citep{ma2025dinfer,wu2025fast,song2025seed,inception2025mercury,gemini_diffusion_2025}, diffusion-style training also enables repeated reuse of the same underlying examples under diverse corruption trajectories, which can extract more information from rare high-quality and long-tail samples and potentially improve overall model capability~\citep{ni2025diffusion,gao2025makes}.
Our work aims to probe this question in the code domain under controlled conditions: given the same data, architecture, and training pipeline as a strong AR baseline, we introduce a block diffusion stage and investigate whether the additional training modes provided by diffusion can translate into tangible gains in code understanding and generation accuracy, rather than merely offering a different decoding mechanism.

\section{Approach}
\subsection{Efficient Knowledge Compression and Training-Inference Alignment in DLLMs}

\textbf{Motivation.} Although random masking can dramatically improve data reusability for diffusion language models \citep{ni2025diffusion,gao2025makes} by repeatedly revisiting the same original examples with different mask patterns applied, it imposes a substantial computational cost. Moreover, some mask patterns only demonstrate negligible training utility, as shown in recent work \citep{kim2025train}.
In particular, many masked tokens are trained in contexts where the correct answer is only weakly constrained, and the resulting gradients are dominated by noisy token co-occurrence rather than a sharp reasoning signal.
Moreover, even when the model compresses knowledge efficiently under some training contexts, this does not automatically translate into good performance under the inference contexts, where a mismatch between the training and inference paradigms has been placed.
We next formalize this effect and use it to motivate practical training curricula.

\subsubsection{Token Reasoning Knowledge under Random Masking}
In the RADD~\citep{ou2025absorbingdiscretediffusionsecretly} formulation, the concrete score at time $t$ for coordinate $i$ takes the below form:
\begin{equation}
s^*(x_t,t)_{(i,\hat x_i)}
\;=\;
\alpha(t)\,
p_0\big(\hat x_i \mid x_t^{\mathrm{UM}}\big),
\end{equation}
where $x_t^{\mathrm{UM}}$ denotes all unmasked tokens at time $t$, and $\alpha(t)$ depends only on $t$ and the forward kernel.
The time dependence is thus factored out, and what remains is the clean-data conditional $p_0(\hat x_i\mid c)$ for various contexts $c$ induced by random masks.

However, not every masked context can lead to a faithful reasoning unmask order towards the final answer. 
Consider a clean sequence:
\begin{equation}
\texttt{a = 1, b = 2, a + b = 3}; 
\texttt{a = 3, b = 4, a + b = 7}.
\end{equation}
If we heavily mask the middle and final parts to obtain:
\begin{equation}
\texttt{a = 1, b = 2, [MASK$_1$] \dots [MASK$_2$] a + b = [MASK$_n$]},\label{eq:badcase}
\end{equation}
the last \texttt{[MASK$_n$]} cannot by itself teach the rule that \texttt{a + b} equals the sum of the preceding \texttt{a} and \texttt{b}, because the model never sees clean evidence for the pair \texttt{a = 3, b = 4}.
From this particular corrupted view, the model mainly learns that some number tends to appear after \texttt{a + b =}, and that certain pairs like \texttt{a = 3, b = 4} and \texttt{7} co-occur. From the clean evidence the model has access to, the correct answer should be 3, yet the masked training signal biases the model toward predicting 7, creating a contradictory and misleading supervision signal.
By contrast, in a sample like below:
\begin{equation}
\texttt{a = 1, b = 2, a + b = 3} 
\quad \;\;\texttt{masked as}\;\;\quad 
\texttt{a = 1, b = 2, a + b = [MASK]} .
\end{equation}
the context strongly constrains the answer. 
In this way, across many such examples, the model can discover a stable mapping from evidence to output, which corresponds to an arithmetic reasoning rule rather than mere memorization.


\begin{definition}[Token Reasoning Knowledge] \label{def:single_knowledge}
Let $x$ be a clean training sequence sampled from the real data distribution $p_0$, and $c$ denote a context extracted from $x$ that is used to predict a single token $x_i$ (similarly for multiple tokens).
For autoregressive training, $c_{\mathrm{AR}} = x_{<i}$, and for diffusion with random masks, $c_t^i = x_t^{\mathrm{UM}}$.
Given a fixed context $c$, the clean-data conditional $p_0(\cdot\mid c)$ induces a \emph{candidate set}:
\begin{equation}
\mathcal{C}(c)
=
\big\{ v \in \mathcal{V} \;:\; p_0(v\mid c) \ge \varepsilon \big\},
\qquad
K(c) = |\mathcal{C}(c)|,\quad \varepsilon > 0.
\end{equation}
We define the \emph{token reasoning knowledge} contained in a training example with context $c$ as the conditional distribution of the next token restricted to its candidate set, denoted compactly by $p_0\big(\mathcal{C}(c)\mid c\big)$
\end{definition}

The model’s objective is to recover this token reasoning knowledge by learning conditionals $p_\theta(x_i^0 \mid c)$ such that, $p_\theta\big(\mathcal{C}(c)\mid c\big)\;\approx\;p_0\big(\mathcal{C}(c)\mid c\big)$.
Concretely, training provides randomly sampled pairs $(c, x_i^0)$ obtained from $(x,i)$ under $p_0$, and the model adjusts $p_\theta(\cdot\mid c)$ based on the empirical distribution of these pairs.
The difficulty of learning this knowledge is analogous to a multi-class classification problem~\citep{dasgupta2025reviewextrememultilabelclassification,kunstner2024heavytailedclassimbalanceadam}: (1) the size of the candidate set $K(c)$; (2) how often each context–label pair $(c, x_i^0)$ appears in the training process.

With respect to $K(c)$, we can distinguish three qualitative regimes:
\begin{itemize}
  \item \textbf{Reasoning regime.}
  $K(c)$ is small and the ground-truth token $x_i^0$ has large probability under $p_0(\cdot\mid c)$.
  The mapping $c\mapsto x_i^0$ is almost deterministic, so repeated samples with similar $c$ provide highly aligned gradients and quickly reinforce a stable reasoning rule.

  \item \textbf{Correlation regime.}
  $K(c)$ is moderate or large, and $x_i^0$ is one of many plausible candidates with comparable probability.
  The model still learns that $x_i^0$ correlates with patterns in $c$, but samples with similar $c$ often yield different targets, so gradients partially cancel and updates primarily fit noisy co-occurrence rather than a sharp rule.

  \item \textbf{Noise regime.}
  $K(c)$ is very large (on the order of $|\mathcal{V}|$) and $p_0(\cdot\mid c)$ is nearly flat.
  The context contains almost no information about $x_i^0$, and the model can only memorize idiosyncratic pairs $(c,x_i^0)$.
\end{itemize}

Because $p_0(\cdot\mid c)$ is defined with respect to the real data distribution, extremely unnatural or noisy contexts $c$ have negligible probability and can be ignored in expectation, as clarified in previous works~\citep{Zhou2023AsymmetricLF,Ghosh2014MakingRM}.
As more relevant evidence is revealed in $c$ (for example, longer and cleaner context), the conditional entropy $H(X_i\mid c)$ cannot increase, 
and the candidate set size $K(c)$ typically shrinks.

For autoregressive architectures and block diffusion with small block sizes, the context $c$ usually contains long, contiguous, and clean left-side evidence, which also matches the inherently autoregressive structure of natural language~\citep{schuurmans2024autoregressive}.
As a result, sampled contexts tend to have relatively small $K(c)$ and allow token reasoning knowledge to be compressed efficiently. 
In contrast, fully bidirectional or large-block diffusion readily produces contexts similar to Eq.~\ref{eq:badcase} that fall into the correlation or noise regimes, dramatically reducing the efficiency of knowledge compression.

\paragraph{\textbf{Context–label Pair $(c, x_i^0)$.}} 
To reliably learn the token reasoning knowledge associated with a context $c$, the model needs a sufficient number of informative context–label pairs $(c, x_i^0)$.
DLLMs based on random masking generate a wide variety of contexts $c$, which greatly increases the number of context–label pairs $(c, x_i^0)$ that the model must learn~\citep{von2025scaling}.
If the total training data is fixed, this diversification reduces the number of effective pairs corresponding to any specific piece of knowledge, lowering its learning efficiency and forcing the model to relearn the same underlying knowledge many more times than an AR model would require~\citep{von2025scaling,sun2025mask,ni2025training}.
Although such random contexts can be viewed as a form of data augmentation, our analysis of the size of $K(c)$ shows that many of these contexts fall into the correlation or noise regimes and therefore cannot be mapped to a clear reasoning rule.
In some cases, such as the pattern in Eq.~\ref{eq:badcase}, the masked context $c$ even encourages learning a wrong association: the model is asked to predict a target under incomplete or misleading evidence.
Such problematic contexts appear frequently when clean evidence is heavily disrupted, for example in purely bidirectional or large-block diffusion settings.

Moreover, the knowledge compressed during training is only useful at inference if it is associated with contexts that actually occur at test time.
Let $\mathcal{C}_{\mathrm{train}}$ denote the set (or distribution) of contexts $c$ sampled during training, and let $\mathcal{C}_{\mathrm{infer}}$ denote the contexts encountered along inference trajectories.
Good performance requires $\mathcal{C}_{\mathrm{train}}$ and $\mathcal{C}_{\mathrm{infer}}$ to be as close as possible. 
For instance, when training with block diffusion of block size $B$ and using left-to-right block-wise decoding of size $B$ at inference, the inference contexts closely match the training contexts, which makes the learned token reasoning knowledge directly applicable at test time.

In summary, to efficiently learn new knowledge while ensuring that the data augmentation induced by DLLM masking remains effective, two conditions should be satisfied:
(1) the model should be exposed to clean and reliable reasoning evidence so that clear reasoning rules can be learned, and
(2) the number of distinct sampled contexts $c$ should not grow excessively, and their form should align as closely as possible with the contexts encountered during inference.
Based on these principles, we next design experiments to explore more suitable training pipelines.

\begin{figure}[t]
    \centering
    \begin{minipage}[t]{0.48\columnwidth}
        \centering
        \includegraphics[width=\linewidth]{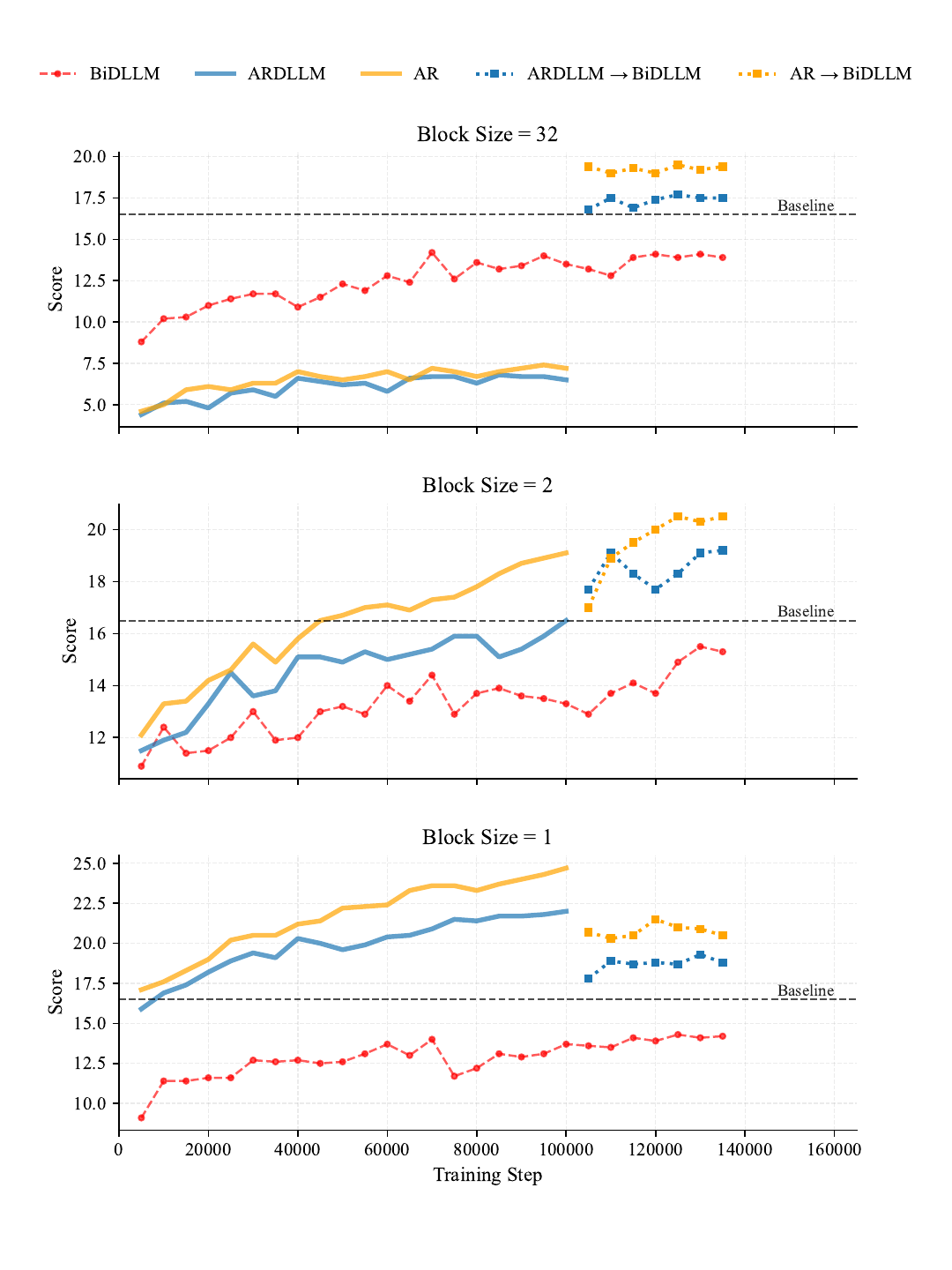}
        \captionof{figure}{
        Training dynamics under different block sizes.
        We compare \textbf{ARDLLM} and \textbf{AR} against a \textbf{BiDLLM} baseline for block sizes 32, 2, and 1 (top to bottom).
        Solid lines indicate training up to 100k steps, while dotted lines denote continued training after switching to BiDLLM.
        The horizontal dashed line marks the fixed baseline reference (0-step AR).
        }
        \label{fig:block-size-ablation}
    \end{minipage}
    \hfill
    \begin{minipage}[t]{0.48\columnwidth}
        \centering
        \includegraphics[width=\linewidth,
        trim=20pt 250pt 90pt 35pt,
        clip]{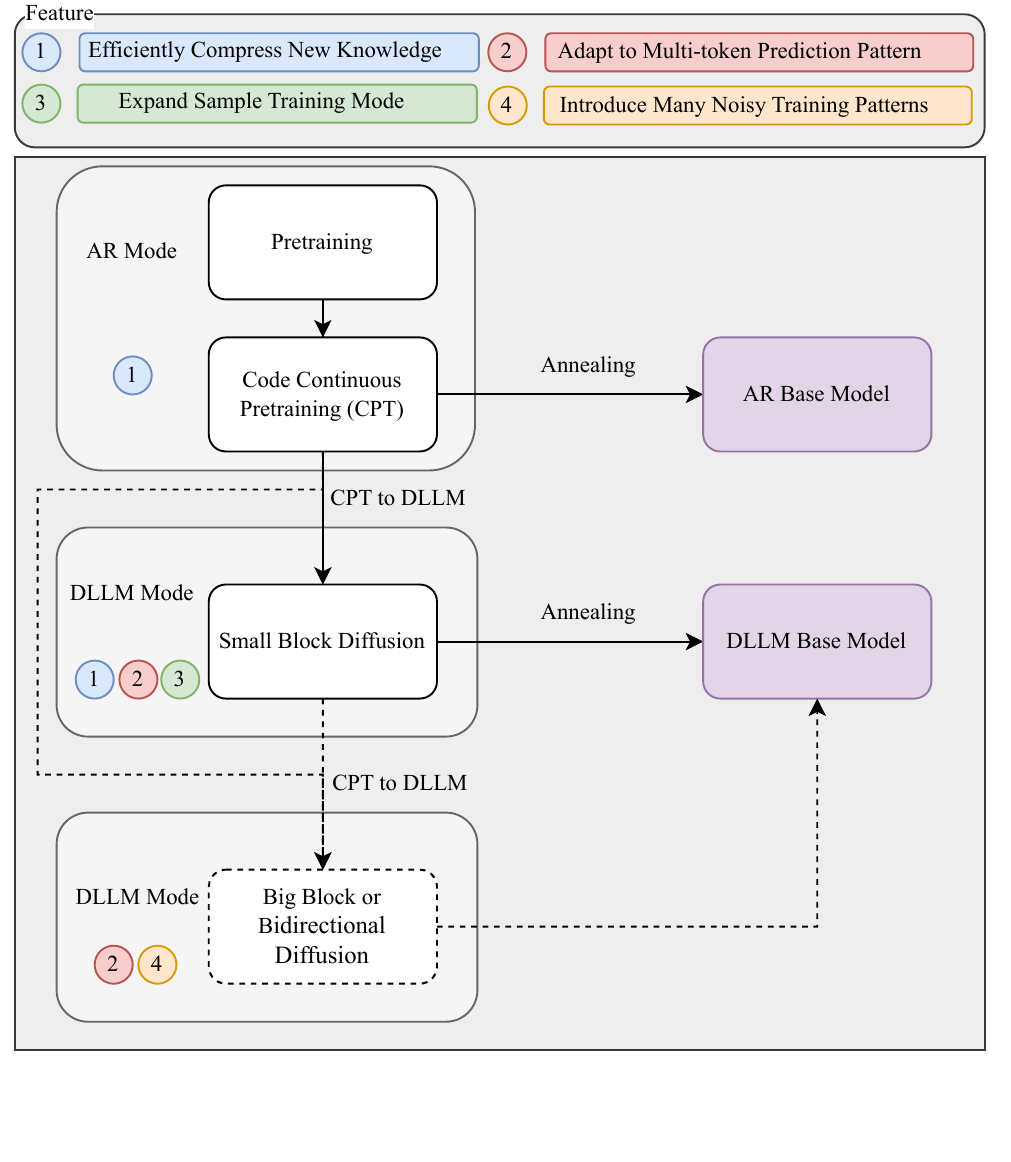}
        \captionof{figure}{
        As shown by the black solid line, we initialize from the pre-annealing checkpoint of Seed-Coder and perform CPT with small-block DLLM to obtain \ourmodel-Base, aiming to study efficient
knowledge acquisition in diffusion-based models. The dashed line denotes an alternative pipeline for larger blocks, where new knowledge is first compressed using an AR model or a small-block DLLM before being transferred to
the large-block diffusion setting.
        }
        \label{fig:pipeline}
    \end{minipage}
\end{figure}

\subsubsection{Curriculum Design and Empirical Results at 2.5B Scale}
Since initializing from an AR checkpoint has become a standard and effective practice, we also start from a 2.5B AR model trained on general-domain data, and use code data as new CPT data.
Starting from this AR checkpoint, we consider the following curricula for learning new knowledge under the same compute budget, and evaluate them using block-wise decoding in the style of LLaDA~\citep{nie2025largelanguagediffusionmodels}:
\begin{enumerate}
  \item[(1)] \textbf{AR $\rightarrow$ BiDLLM}: continue pure AR training on the new data, then perform CPT to a bidirectional DLLM;
  \item[(2)] \textbf{ARDLLM $\rightarrow$ BiDLLM}: continue training with a causal-structured DLLM (AR-style diffusion), then CPT to a bidirectional DLLM;
  \item[(3)] \textbf{BiDLLM}: directly CPT the AR checkpoint into a bidirectional DLLM and train new knowledge in that regime.
\end{enumerate}
As shown in Fig.~\ref{fig:block-size-ablation}, we report the average
performance across multiple code benchmarks. Under the same compute budget and after continued pre-training (CPT),
the overall performance follows the ranking $\text{(1)} \;>\; \text{(2)} \;>\; \text{(3)}\quad\text{under the same compute.}$

Scheme~(1) preserves a purely AR training structure before CPT.
Under small-block decoding (block size 1 or 2), its $\mathcal{C}_{\mathrm{infer}}$ are almost identical to its $\mathcal{C}_{\mathrm{train}}$ and satisfy the two principles from the previous section: the model sees clean reasoning evidence, and the number of distinct contexts remains controlled and well aligned with small-block inference.
Consequently, before CPT to BiDLLM, scheme~(1) consistently achieves the best performance for small blocks.
For large-block decoding (block size 32), however, scheme~(1) underperforms scheme~(3), because the training context distribution $\mathcal{C}_{\mathrm{train}}$ of pure AR does not cover the bidirectional patterns required by large-block decoding.
Formally, with causal attention, tokens later in the block cannot influence earlier ones, so when decoding an entire block at once, the generation process becomes less stable.

After CPT into a bidirectional DLLM, scheme~(1) performs well across all block sizes.
This suggests that the AR phase has already compressed the new token reasoning knowledge effectively in a well-aligned context family for small blocks, and the subsequent BiDLLM CPT primarily adapts the model to the larger-block context distribution.

It is worth noting that CPT into a fully bidirectional model introduces some degradation for block-1 decoding: both pure AR and AR$\rightarrow$BiDLLM lose a similar amount of performance between 0 and 100k CPT steps at block size 1. However, the results also show that scheme~(1) achieves higher learning efficiency during the AR phase, so even after the CPT-induced drop, it still attains better final performance.

Interestingly, scheme~(2) also achieves strong performance after CPT.
In particular, after converting to BiDLLM, it outperforms scheme~(3) under block-32 decoding, despite having seen a similar amount (or even less) clean evidence during training.
One explanation is that the training context distribution $\mathcal{C}_{\mathrm{train}}$ for AR-style diffusion in scheme~(2) more closely matches the prompt–response pattern commonly used at inference, where most reasoning evidence lies on the left.
Thus scheme~(2) enjoys better training–inference alignment than scheme~(3), and this advantage persists even when both are eventually converted to a fully bidirectional DLLM.

Finally, based on the above theoretical analysis and empirical results, we recommend the following training procedure:
\begin{enumerate}
    \item For new knowledge, first use AR training to efficiently compress it;
    \item Then perform CPT with a small-block diffusion objective, leveraging its data-augmentation properties to further improve model quality; 
    \item if one wishes to explore larger block diffusion, additional CPT can be applied starting from the model obtained in step~(2).
\end{enumerate}
In our final system, we adopt steps~(1) and~(2) to train \ourmodel, as illustrated by the solid path leading to the ``DLLM Base Model'' in Fig.~\ref{fig:pipeline}.

\subsection{Warmup for Stable DLLM Continued Pretraining}
Previous work~\citep{gong2024scaling} has reported that CPT of mask diffusion language models (DLLMs) is highly sensitive to the learning rate.
In practice, the stability of the loss and gradient norm is also affected by architecture (e.g., attention pattern, logit parametrization), numerical precision, and infrastructure details.
This motivates a more robust warmup procedure for DLLM CPT.

To reduce the architectural and objective gap between AR and DLLM, we reuse the AR token head and logit-shift parametrization, and only change the attention pattern from a causal lower-triangular mask to a bidirectional, full attention mask.
Even under this minimal change, naive AR$\rightarrow$DLLM CPT shows a high initial loss and large gradient-norm spikes.

We attribute the instability mainly to three factors:
(i) the change of attention mask, which induces a structural distribution shift in internal representations;
(ii) the higher task difficulty when the corruption process masks a large fraction of tokens, compared to AR next-token prediction;
(iii) the ELBO-motivated loss weight $w(t)$ in the DLLM objective (Eq.~\ref{eq:dlm-loss}),
where $w(t)$ can be large at low masking ratios (e.g., under a linear noise schedule, masking $10\%$ of tokens yields $w(t)\approx 10$). This effectively acts as loss scaling that amplifies gradient norms, making training less stable.

Rather than annealing the attention mask~\citep{gong2024scaling} (which is inconvenient for highly optimized kernels such as FlashAttention that assume a fixed mask), we warm up only the corruption process.
In standard DLLM, the corruption level $t$ is sampled uniformly from $[0,1]$, and the masking ratio ranges from $0$ to $1$.
During warmup, we cap the maximum corruption level: $t \sim \mathcal{U}(0, u_{\max}),$
and linearly increase $q_{\max}$ from a small value $q_{\text{init}}$ (e.g., $10^{-3}$) to $1$ over $S_{\text{warmup}}$ steps:
\[
u_{\max}(s) = u_{\text{init}} + (1 - u_{\text{init}})\cdot \frac{s}{S_{\text{warmup}}}, \quad s=0,\dots,S_{\text{warmup}}.
\]
This implements a curriculum from easy (low mask ratio, almost AR-like) reconstruction to the full DLLM regime. To further suppress gradient spikes, we drop $w(t)$ in the warmup phase and optimize
\begin{equation}
\mathcal{L}^{\text{warmup}}_{\mathrm{DLLM}}(\theta) = -\mathbb{E}_{\mathbf{x}^0\sim p_{\mathrm{data}},\, t\sim \mathcal{U}(0,u_{\max}),\, \mathbf{x}^t\sim q(\mathbf{x}^t\mid \mathbf{x}^0)}
\Big[\sum_{i=1}^N   {1}[\mathbf{x}^t_{i}=\text{MASK}]\,\log p_\theta(\mathbf{x}^0_i\mid \mathbf{x}^t_{1:N} )\Big].
\label{eq:dlm-loss-warmup}
\end{equation}
After warmup, we revert to the original loss in Eq.~\eqref{eq:dlm-loss} with $t\sim\mathcal{U}(0,1)$.

\begin{figure}[t]
    \centering
    \begin{subfigure}{0.32\linewidth}
        \centering
        \includegraphics[width=\linewidth]{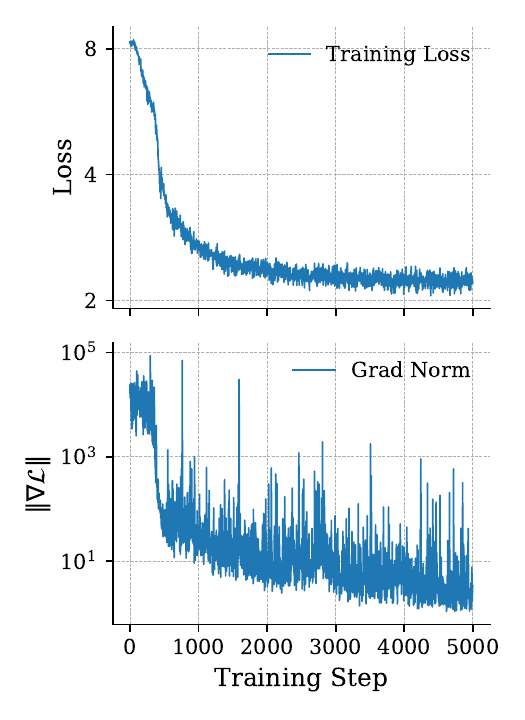}
        \caption*{BiDLLM Without Warmup}
    \end{subfigure}
    \hfill
    \begin{subfigure}{0.32\linewidth}
        \centering
        \includegraphics[width=\linewidth]{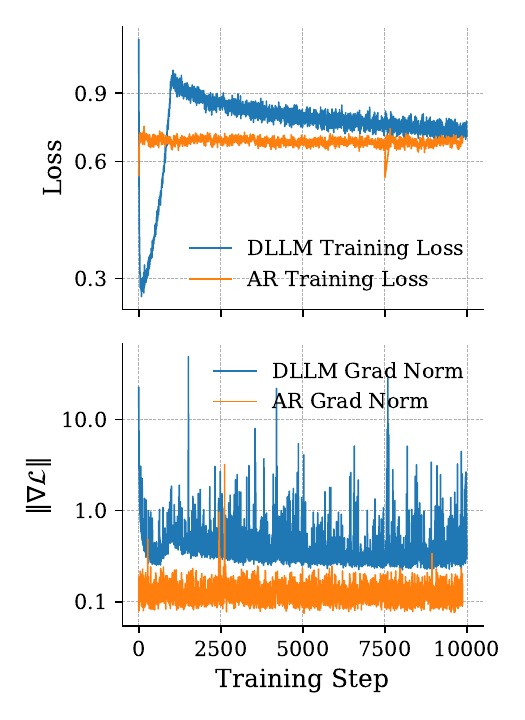}
        \caption*{BiDLLM With Warmup}
    \end{subfigure}
    \hfill
    \begin{subfigure}{0.32\linewidth}
        \centering
        \includegraphics[width=\linewidth]{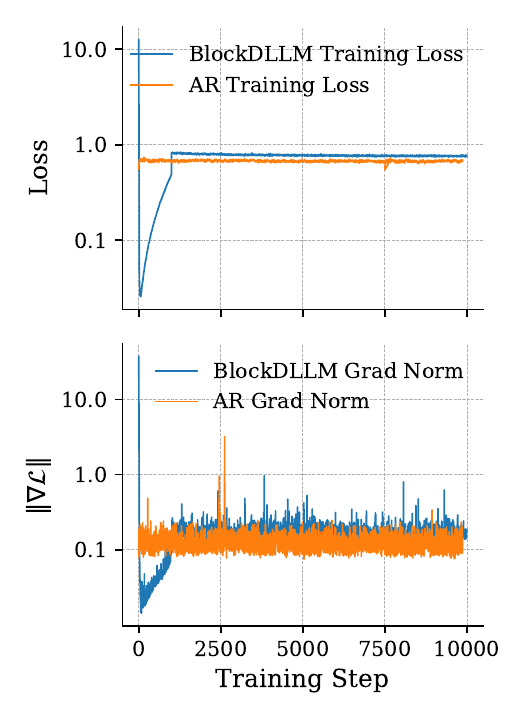}
        \caption*{BlockDLLM With Warmup (no-shift)}
    \end{subfigure}

    \caption{\textbf{Comparison of training stability before and after applying warmup.}
    The left figure shows the behavior without warmup, and the right figure shows the behavior with warmup.
    When warmup is used, both the training loss and gradient norm become significantly more stable, and they quickly decrease to a level comparable to that of the AR continual pretraining stage. BiDLLM refers to a purely bidirectional masked diffusion model, BlockDLLM to a block-masked diffusion model, and no-shift indicates that token logit shifting is disabled, which is the configuration adopted in our final model.}
    \label{fig:warmup_comparison}
\end{figure}

This warmup produces a much smoother CPT trajectory: the gradient norm spike at the AR$\rightarrow$DLLM boundary is strongly reduced, and the loss follows a characteristic ``$\sqrt{\cdot}$-shaped'' curve (first decreasing on easy tasks, then increasing as the mask ratio grows, and finally decreasing again).
The peak loss during warmup is significantly lower than the naive AR$\rightarrow$DLLM baseline, and the final loss is comparable to AR$\rightarrow$AR CPT.
We observe similarly stable behavior when we remove the logit shift and train a block-diffusion DLLM with the same warmup (see Fig.~\ref{fig:warmup_comparison}).

\subsection{Block-wise Clipped Noise Scheduling for Masked Block Diffusion}
\label{subsec:block-quantized-schedule}

\paragraph{Motivation.}
In masked \emph{block} diffusion training, at each step we corrupt only a single contiguous block of tokens rather than the full sequence.
If one reuses a \emph{global} continuous-time schedule $t \!\in\! [0,1] \mapsto u(t)$ (with $q$ the corruption/mask rate) that was designed for whole-sequence diffusion, a significant portion of training steps produce weak or even zero learning signal when the block length $B$ is small.
Concretely, when the corruption is applied only inside a block $\mathcal{B}$ of size $B$, the expected number of corrupted tokens at time $t$ is $\mathbb{E}[m \mid t] = B\,u(t)$ and the probability of observing \emph{no} corrupted token is $\Pr[m=0 \mid t] \;=\; (1-u(t))^{B}. \label{eq:no-loss-prob}$.
Under a standard global linear schedule $u(t)=1-t$ with $t\sim\mathrm{Unif}[0,1]$, the fraction of steps with $m=0$ equals
\begin{equation}
\mathbb{E}_{t}[(1-t)^{B}] \;=\; \int_{0}^{1} (1-t)^{B}\,dt \;=\; \frac{1}{B+1},\label{eq:zero-fraction-global}
\end{equation}
which is non-negligible for typical block sizes (e.g., $1/3$ for $B{=}2$, $1/5$ for $B{=}4$, $1/9$ for $B{=}8$).

Instead of redesigning the global schedule, we adopt a simple block-aware sampling rule.
At each training step, after sampling $t$, we define a block-specific mask rate by clipping
$u_{\mathrm{blk}}(t)\;=\;\min\bigl(1,\; \max(u(t),\,1/B)\bigr),\label{eq:block-schedule}$
so that $q_{\mathrm{blk}}(t) \in [1/B,\,1]$ for all $t$.
This guarantees that the expected number of masked tokens in the block satisfies $\mathbb{E}[m \mid t] = B\,u_{\mathrm{blk}}(t) \ge 1$. In addition, it prevents the loss weight $w(t)=\tfrac{1}{u_{\mathrm{blk}}(t)}$ for those tokens from becoming excessively large, thereby promoting more stable training.

Given $u_{\mathrm{blk}}(t)$, we independently mask each token in the chosen block $\mathcal{B}$ with probability $u_{\mathrm{blk}}(t)$, while tokens outside $\mathcal{B}$ remain clean and provide context.
To further ensure that every block contributes to the loss, we apply the following fallback rule:
if, after sampling, no token in $\mathcal{B}$ is masked (i.e., $m=0$), we uniformly sample one position in $\mathcal{B}$ and force it to be masked.
In this way, every training step contains at least one supervised token inside the block, while still preserving the overall shape of the original schedule $u(t)$.

\section{Experiments}
\label{sec:experiments}
We conduct extensive experiments to compare the \emph{base} model and its instruction-tuned (\emph{instruction}) variant across a diverse set of code-centric benchmarks. 

\subsection{Experiments Setting}
To preserve representation plasticity and enable structural adaptation to the target data distribution, we perform continuous pretraining from a Seed-Coder pre-annealing checkpoint, using a context length of 8192 with packed sequences. We reuse the Seed-Coder training pipeline and compress its multi-stage continuous pretraining data to a total of 1.3T tokens via subsampling. The supervised fine-tuning (SFT) stage fully reuses the original Seed-Coder SFT dataset.
We adopt block diffusion with a block size of 4 for this continuous pretraining stage. Since the no-logit-shift formulation is more consistent with the absorbing diffusion paradigm, where each masked position predicts itself and the input and prediction targets are aligned at both the token and sentence levels, we follow a no-logit-shift design similar to that used in LLADA and SDAR.

To prevent repeated compilation of flex attention, we make the attention between packed samples mutually visible, so that the same attention mask and operator can be reused in every forward pass.
To maintain training efficiency, we adopt the same packing strategy during SFT. However, after each sample we randomly append 1–4 `<eos>` tokens, enabling the model to preserve the ability to generate variable-length outputs within each packed block.

In our experiments, we compare our method against a broad set of strong
AR models and DLLM
that have demonstrated competitive performance on code generation tasks.

The AR baselines include
StarCoder2~\citep{lozhkov2024starcoder2}, DeepSeek-Coder~\citep{deepseek_coder}, CodeQwen1.5~\citep{codeqwen1.5}, OpenCoder~\citep{huang2025opencoder}, Qwen2.5-Coder~\citep{hui2024qwen2},
Seed-Coder~\citep{seed2025seed}, CodeLlama~\citep{codellama}, and Llama3.1~\citep{meta2024llama31}.

The DLLM baselines include
LLaDA~\citep{nie2025largelanguagediffusionmodels}, Dream~\citep{dream2025}, DiffuCoder~\citep{gong2025diffucoder}, Dream-Coder~\citep{xie2025dream}, LLaDA-MoE~\citep{zhu2025llada}, Fast-dLLMv2~\citep{wu2025fast}, SDAR~\citep{cheng2025sdar},
Seed-Diffusion-Preview, LLaDA2.0~\citep{bie2025llada2}, SDLM~\citep{liu2025sequential}, WeDLM~\citep{liu2025wedlmreconcilingdiffusionlanguage}, Mercury Coder~\citep{inception2025mercury}, and Gemini Diffusion~\citep{gemini_diffusion_2025}.
\begin{table}[ht]
\centering
\resizebox{0.75\textwidth}{!}{
\begin{tabular}{lcrcccc}
\toprule
\multirow{2}{*}{\textbf{Model}}& \multicolumn{1}{c}{\multirow{2}{*}{\textbf{Type}}} & \multicolumn{1}{r|}{\multirow{2}{*}{\textbf{Size}}} & \multicolumn{2}{c|}{\textbf{HumanEval}} & \multicolumn{2}{c}{\textbf{MBPP}} \\
 && \multicolumn{1}{r|}{} & \textit{HE} & \multicolumn{1}{c|}{\textit{HE$^+$}} & \textit{MBPP} & \textit{MBPP$^+$} \\ \midrule
\multicolumn{7}{c}{\textasciitilde 8B Models} \\ \midrule
StarCoder2-7B&AR & \multicolumn{1}{r|}{7B} & 35.4 & \multicolumn{1}{c|}{29.9} & 54.4 & 45.6 \\
DeepSeek-Coder-6.7B-Base&AR & \multicolumn{1}{r|}{6.7B} & 47.6 & \multicolumn{1}{c|}{39.6} & 70.2 & 56.6 \\
CodeQwen1.5-7B&AR & \multicolumn{1}{r|}{7B} & 51.8 & \multicolumn{1}{c|}{45.7} & 72.2 & 60.2 \\
OpenCoder-8B-Base&AR & \multicolumn{1}{r|}{8B} & 66.5 & \multicolumn{1}{c|}{63.4} & 79.9 & \fst{70.4} \\
Qwen2.5-Coder-7B&AR & \multicolumn{1}{r|}{7B} & 72.0 & \multicolumn{1}{c|}{67.1} & 79.4 & 68.3 \\
Seed-Coder-8B-Base&AR & \multicolumn{1}{r|}{8B} & 77.4 & \multicolumn{1}{c|}{68.3} & 82.0 & 69.0 \\ 
LLaDA-8B-Base&DLLM & \multicolumn{1}{r|}{8B} & 35.4 & \multicolumn{1}{c|}{30.5} & 50.1 & 42.1 \\
Dream-7B-Base&DLLM & \multicolumn{1}{r|}{7B} & 56.7 & \multicolumn{1}{c|}{50.0} & 68.7 & 57.4 \\
DiffuCoder-7B-Base&DLLM & \multicolumn{1}{r|}{7B} & 67.1 & \multicolumn{1}{c|}{60.4} & 74.2 & 60.9 \\
Dream-Coder-7B-Base&DLLM & \multicolumn{1}{r|}{7B} & 66.5 & \multicolumn{1}{c|}{60.4} & 75.9 & 61.6 \\
LLaDA-MoE-7B-A1B-Base&DLLM & \multicolumn{1}{r|}{1B/7B} & 45.7 & \multicolumn{1}{c|}{-} & 52.4 & - \\
WeDLM-8B-Base&DLLM & \multicolumn{1}{r|}{8B} & 75.0 & \multicolumn{1}{c|}{68.9} & 67.0 & - \\
\graybg \ourmodel-8B-Base&DLLM & \multicolumn{1}{r|}{8B} & \fst{79.3} & \multicolumn{1}{c|}{\fst{73.8}} & \fst{83.6} & 67.7 \\
\midrule
\multicolumn{7}{c}{\grayt 13B+ Models} \\ \midrule
\grayt StarCoder2-15B&AR & \multicolumn{1}{r|}{\grayt 15B} & \grayt 46.3 & \multicolumn{1}{c|}{\grayt 37.8} & \grayt 66.2 & \grayt 53.1 \\
\grayt CodeLlama-70B-Base&AR & \multicolumn{1}{r|}{\grayt 70B} & \grayt 52.4 & \multicolumn{1}{c|}{\grayt 50.6} & \grayt 71.0 & \grayt 65.6 \\
\grayt Llama-3.1-70B-Base&AR & \multicolumn{1}{r|}{\grayt 70B} & \grayt 54.9 & \multicolumn{1}{c|}{\grayt 51.2} & \grayt 81.4 & \grayt 67.2 \\
\grayt DeepSeek-Coder-33B-Base&AR & \multicolumn{1}{r|}{\grayt 33B} & \grayt 54.9 & \multicolumn{1}{c|}{\grayt 47.6} & \grayt 74.2 & \grayt 60.7 \\
\grayt DeepSeek-Coder-V2-Lite-Base&AR & \multicolumn{1}{r|}{\grayt 2.4B/16B} & \grayt 40.9 & \multicolumn{1}{c|}{\grayt 34.1} & \grayt 71.9 & \grayt 59.4 \\
\grayt Qwen2.5-Coder-14B&AR & \multicolumn{1}{r|}{\grayt 14B} & \grayt \fst{83.5} & \multicolumn{1}{c|}{\grayt \fst{75.6}} & \grayt \fst{83.6} & \grayt \fst{69.8} \\
\bottomrule
\end{tabular}
}
\caption{Performance of various base models on HumanEval($^+$) and MBPP($^+$).}
\label{tab:base_humaneval_mbpp}
\end{table}
\begin{table}[ht]
\centering
\resizebox{\textwidth}{!}{
\begin{tabular}{lrcccccccc|c}
\toprule
\textbf{Model} & \multicolumn{1}{r|}{\textbf{Size}} & Python & C++ & Java & PHP & TS & C\# & Bash & JS & \textbf{Average} \\ \midrule
\multicolumn{11}{c}{\textasciitilde 8B Models} \\ \midrule
StarCoder2-7B & \multicolumn{1}{r|}{7B} & 35.4 & 40.4 & 38.0 & 30.4 & 34.0 & 46.2 & 13.9 & 36.0 & 34.3 \\
DeepSeek-Coder-6.7B-Base & \multicolumn{1}{r|}{6.7B} & 49.4 & 50.3 & 43.0 & 38.5 & 49.7 & 50.0 & 28.5 & 48.4 & 44.7 \\
CodeQwen1.5-7B & \multicolumn{1}{r|}{7B} & 51.8 & 52.2 & 42.4 & 46.6 & 52.2 & 55.7 & 36.7 & 49.7 & 48.4 \\
OpenCoder-8B-Base & \multicolumn{1}{r|}{8B} & 66.5 & 63.4 & 63.9 & 61.5 & 68.6 & 54.3 & 44.3 & 65.8 & 61.0 \\
Qwen2.5-Coder-7B & \multicolumn{1}{r|}{7B} & 72.0 & 62.1 & 53.2 & 59.0 & 64.2 & \fst{60.8} & 38.6 & 60.3 & 58.8 \\
Seed-Coder-8B-Base & \multicolumn{1}{r|}{8B} & 77.4 & \fst{69.6} & 72.8 & 63.9 & \fst{77.4} & 53.8 & 48.1 & \fst{77.6} & 67.6 \\ 
\graybg \ourmodel-8B-Base & \multicolumn{1}{r|}{8B} & \fst{80.5} & 69.4 & \fst{74.1} & \fst{74.4} & 74.8 & \fst{70.3} & \fst{53.2} & 73.1 & \fst{71.2} \\ 
\midrule
\multicolumn{11}{c}{\grayt 13B+ Models} \\ \midrule
\grayt StarCoder2-15B & \multicolumn{1}{r|}{\grayt 15B} & \grayt 46.3 & \grayt 47.2 & \grayt 46.2 & \grayt 39.1 & \grayt 42.1 & \grayt 53.2 & \grayt 15.8 & \grayt 43.5 & \grayt 41.7 \\
\grayt CodeLlama-70B-Base & \multicolumn{1}{r|}{\grayt 70B} & \grayt 52.4 & \grayt 49.7 & \grayt 44.7 & \grayt 46.6 & \grayt 57.2 & \grayt 46.7 & \grayt 31.6 & \grayt 56.5 & \grayt 48.2 \\
\grayt Llama-3.1-70B-Base & \multicolumn{1}{r|}{\grayt 70B} & \grayt 54.9 & \grayt 41.0 & \grayt 41.1 & \grayt 48.4 & \grayt 57.9 & \grayt 44.2 & \grayt 29.1 & \grayt 55.3 & \grayt 46.5  \\
\grayt DeepSeek-Coder-33B-Base & \multicolumn{1}{r|}{\grayt 33B} & \grayt 56.1 & \grayt 58.4 & \grayt \fst{51.9} & \grayt 44.1 & \grayt 52.8 & \grayt 51.3 & \grayt 32.3 & \grayt 55.3 & \grayt 50.3 \\
\grayt DeepSeek-Coder-V2-Lite-Base & \multicolumn{1}{r|}{\grayt 2.4B/16B} & \grayt 40.9 & \grayt 45.9 & \grayt 34.8 & \grayt 47.2 & \grayt 48.4 & \grayt 41.7 & \grayt 19.6 & \grayt 44.7 & \grayt 40.4 \\ 
\grayt Qwen2.5-Coder-14B & \multicolumn{1}{r|}{\grayt 14B} & \grayt \fst{83.5} & \grayt \fst{69.6} & \grayt 46.8 & \grayt \fst{64.6} & \grayt \fst{69.2} & \grayt \fst{63.3} & \grayt \fst{39.9} & \grayt \fst{61.5} & \grayt \fst{62.3} \\
\bottomrule
\end{tabular}
}
\caption{Performance of base models on MultiPL-E.}
\label{tab:base_multipl-e}
\end{table}
\begin{table}[ht]
\centering
\resizebox{0.65\textwidth}{!}{
\begin{tabular}{lrcc}
\toprule
\multirow{2}{*}{\textbf{Model}} & \multicolumn{1}{r|}{\multirow{2}{*}{\textbf{Size}}} & \multicolumn{2}{c}{\textbf{CRUXEval}} \\
    & \multicolumn{1}{r|}{} & \textit{Input-CoT} & \textit{Output-CoT} \\ \midrule
\multicolumn{4}{c}{\textasciitilde 8B Models} \\ \midrule
StarCoder2-7B & \multicolumn{1}{r|}{7B} & 39.5 & 35.1 \\
DeepSeek-Coder-6.7B-Base & \multicolumn{1}{r|}{6.7B} & 39.0 & 41.0 \\
OpenCoder-8B-Base & \multicolumn{1}{r|}{8B} & 43.3 & 43.9 \\
Qwen2.5-Coder-7B & \multicolumn{1}{r|}{7B} & \fst{56.5} & 56.0 \\
Seed-Coder-8B-Base & \multicolumn{1}{r|}{8B} & 52.0 & 54.8 \\
\graybg \ourmodel-8B-Base & \multicolumn{1}{r|}{8B} & 53.8 & \fst{60.0} \\\midrule
\multicolumn{4}{c}{\grayt 13B+ Models} \\ \midrule
\grayt StarCoder2-15B & \multicolumn{1}{r|}{\grayt 15B} & \grayt 46.1 & \grayt 47.6 \\
\grayt CodeLlama-34B-Base & \multicolumn{1}{r|}{\grayt 34B} & \grayt 49.4 & \grayt 43.9 \\
\grayt DeepSeek-Coder-33B-Base & \multicolumn{1}{r|}{\grayt 33B} & \grayt 50.6 & \grayt 48.8 \\
\grayt DeepSeek-Coder-V2-Lite-Base & \multicolumn{1}{r|}{\grayt 2.4B/16B} & \grayt 53.4 & \grayt 46.1 \\
\grayt Qwen2.5-Coder-14B & \multicolumn{1}{r|}{\grayt 14B} & \grayt \fst{60.6} & \grayt \fst{66.4} \\
\bottomrule
\end{tabular}
}
\caption{Performance of base models CRUXEval.}
\label{tab:base_bigcodebench_cruxeval}
\end{table}

\begin{table}[ht]
\centering
\resizebox{0.77\textwidth}{!}{
\begin{tabular}{lcrcccc}
\toprule
\multirow{2}{*}{\textbf{Model}} & \multicolumn{1}{c}{\multirow{2}{*}{\textbf{Type}}} &\multicolumn{1}{r|}{\multirow{2}{*}{\textbf{Size}}} & \multicolumn{2}{c|}{\textbf{HumanEval}} & \multicolumn{2}{c}{\textbf{MBPP}} \\
& & \multicolumn{1}{r|}{} & \textit{HE} & \multicolumn{1}{c|}{\textit{HE$^+$}} & \textit{MBPP} & \textit{MBPP$^+$} \\ \midrule
\multicolumn{6}{c}{\textasciitilde 8B Models} \\ \midrule
CodeLlama-7B-Instruct &AR & \multicolumn{1}{r|}{7B} & 40.9 & \multicolumn{1}{c|}{33.5} & 54.0 & 44.4 \\
DeepSeek-Coder-6.7B-Instruct &AR & \multicolumn{1}{r|}{6.7B} & \multicolumn{1}{c}{74.4} & \multicolumn{1}{c|}{71.3} & \multicolumn{1}{c}{74.9} & 65.6 \\
CodeQwen1.5-7B-Chat &AR & \multicolumn{1}{r|}{7B} & \multicolumn{1}{c}{83.5} & \multicolumn{1}{c|}{78.7} & \multicolumn{1}{c}{77.7} & 67.2\\
Yi-Coder-9B-Chat &AR & \multicolumn{1}{r|}{9B} & \multicolumn{1}{c}{82.3} & \multicolumn{1}{c|}{74.4} & \multicolumn{1}{c}{82.0} & 69.0 \\
Llama-3.1-8B-Instruct &AR & \multicolumn{1}{r|}{8B} & \multicolumn{1}{c}{68.3} & \multicolumn{1}{c|}{59.8} & \multicolumn{1}{c}{70.1} & 59.0 \\
OpenCoder-8B-Instruct &AR & \multicolumn{1}{r|}{8B} & \multicolumn{1}{c}{83.5} & \multicolumn{1}{c|}{78.7} & \multicolumn{1}{c}{79.1} & 69.0 \\
Qwen2.5-Coder-7B-Instruct &AR & \multicolumn{1}{r|}{7B} & \multicolumn{1}{c}{\fst{88.4}} & \multicolumn{1}{c|}{\fst{84.1}} & \multicolumn{1}{c}{83.5} & 71.7 \\
Qwen3-8B &AR & \multicolumn{1}{r|}{8B} & \multicolumn{1}{c}{84.8} & \multicolumn{1}{c|}{80.5} & \multicolumn{1}{c}{77.0} & 67.2 \\
Seed-Coder-8B-Instruct &AR & \multicolumn{1}{r|}{8B} & \multicolumn{1}{c}{84.8} & \multicolumn{1}{c|}{78.7} & 85.2 & 71.2\\ 
LLaDA-8B-Instruct &DLLM & \multicolumn{1}{r|}{8B} & \multicolumn{1}{c}{49.4} & \multicolumn{1}{c|}{-} & 41.0 & -\\ 
Dream-7B-Instruct &DLLM & \multicolumn{1}{r|}{7B} & \multicolumn{1}{c}{63.4} & \multicolumn{1}{c|}{-} & 68.3 & -\\ 
LLaDA-MoE-7B-Instruct &DLLM & \multicolumn{1}{r|}{1B/7B} & \multicolumn{1}{c}{61.6} & \multicolumn{1}{c|}{-} & 70.0 & -\\ 
Fast-dLLMv2 &DLLM & \multicolumn{1}{r|}{7B} & \multicolumn{1}{c}{43.9} & \multicolumn{1}{c|}{40.2} & 50.0 & 41.3\\ 
DiffuCoder-7B-Instruct &DLLM & \multicolumn{1}{r|}{7B} & \multicolumn{1}{c}{72.0} & \multicolumn{1}{c|}{65.2} & 75.1 & 61.9\\
Dream-Coder-7B-Instruct &DLLM & \multicolumn{1}{r|}{7B} & \multicolumn{1}{c}{82.9} & \multicolumn{1}{c|}{-} & 79.6 & -\\
SDAR-8B-Chat &DLLM & \multicolumn{1}{r|}{8B} & \multicolumn{1}{c}{78.7} & \multicolumn{1}{c|}{-} & 72.0 & -\\ 
WeDLM-8B-Chat &DLLM & \multicolumn{1}{r|}{8B} & \multicolumn{1}{c}{80.5} & \multicolumn{1}{c|}{73.8} & 70.5 & -\\ 
\graybg \ourmodel-8B-Instruct &DLLM & \multicolumn{1}{r|}{8B} & \multicolumn{1}{c}{86.6} & \multicolumn{1}{c|}{82.3} & \fst{85.7} & \fst{72.8}\\ 
\midrule
\multicolumn{6}{c}{\grayt 13B+ Models and External API} \\ \midrule
\grayt StarCoder2-15B-Instruct &AR & \multicolumn{1}{r|}{\grayt 15B} & \multicolumn{1}{c}{\grayt 67.7} & \multicolumn{1}{c|}{\grayt 60.4} & \multicolumn{1}{c}{\grayt 78.0} & \grayt 65.1 \\
\grayt Codestral-22B &AR & \multicolumn{1}{r|}{\grayt 22B} & \multicolumn{1}{c}{\grayt 81.1} & \multicolumn{1}{c|}{\grayt 73.2} & \multicolumn{1}{c}{\grayt 78.2} & \grayt 62.2 \\
\grayt CodeLlama-70B-Instruct &AR & \multicolumn{1}{r|}{\grayt 70B} & \multicolumn{1}{c}{\grayt 72.0} & \multicolumn{1}{c|}{\grayt 65.9} & \multicolumn{1}{c}{\grayt 77.8} & \grayt 64.6 \\
\grayt DeepSeek-Coder-33B-Instruct &AR & \multicolumn{1}{r|}{\grayt 33B} & \multicolumn{1}{c}{\grayt 81.1} & \multicolumn{1}{c|}{\grayt 75.0} & \multicolumn{1}{c}{\grayt 80.4} & \grayt 70.1  \\ 
\grayt DeepSeek-Coder-V2-Lite-Istruct &AR & \multicolumn{1}{r|}{\grayt 2.4B/16B} & \multicolumn{1}{c}{\grayt 81.1} & \multicolumn{1}{c|}{\grayt 75.6} & \multicolumn{1}{c}{\grayt 82.8} & \grayt 70.4 \\
\grayt DeepSeek-Coder-V2-Instruct &AR & \multicolumn{1}{r|}{\grayt 21B/236B} & \multicolumn{1}{c}{\grayt 85.4} & \multicolumn{1}{c|}{\grayt 82.3} & \multicolumn{1}{c}{\grayt 89.4} & \grayt 75.1 \\
\grayt Qwen2.5-Coder-14B-Instruct &AR & \multicolumn{1}{r|}{\grayt 14B} & \multicolumn{1}{c}{\grayt 89.6} & \multicolumn{1}{c|}{\grayt 87.2} & \multicolumn{1}{c}{\grayt 86.2} & \grayt 72.8 \\
\grayt Qwen2.5-Coder-32B-Instruct &AR & \multicolumn{1}{r|}{\grayt 32B} & \multicolumn{1}{c}{\grayt 92.7} & \multicolumn{1}{c|}{\grayt \fst{87.2}} & \multicolumn{1}{c}{\grayt \fst{90.2}} & \grayt 75.1 \\
SDAR-30B-Chat &DLLM & \multicolumn{1}{r|}{3B/30B} & \multicolumn{1}{c}{87.2} & \multicolumn{1}{c|}{-} & 71.6 & -\\
LLaDA2.0-mini &DLLM & \multicolumn{1}{r|}{1B/16B} & \multicolumn{1}{c}{86.6} & \multicolumn{1}{c|}{79.9} & 81.5 & 74.1\\ 
LLaDA2.0-flash &DLLM & \multicolumn{1}{r|}{6B/100B} & \multicolumn{1}{c}{\fst{94.5}} & \multicolumn{1}{c|}{\fst{87.8}} & 88.3 & \fst{79.6}\\ 
SDLM-32B &DLLM & \multicolumn{1}{r|}{32B} & \multicolumn{1}{c}{81.1} & \multicolumn{1}{c|}{73.8} & 80.9 & 57.1\\ 
Seed-Diffusion-Preview(0705) &DLLM & \multicolumn{1}{r|}{-} & \multicolumn{1}{c}{82.8} & \multicolumn{1}{c|}{-} & 79.4 & -\\ 
Mercury Coder&DLLM & \multicolumn{1}{r|}{-} & \multicolumn{1}{c}{90.0} & \multicolumn{1}{c|}{-} & 77.1 & -\\ 
Gemini Diffusion &DLLM & \multicolumn{1}{r|}{-} & \multicolumn{1}{c}{89.6} & \multicolumn{1}{c|}{-} & 76.0 & -\\ 
\bottomrule
\end{tabular}
}
\caption{Performance of various base models on HumanEval($^+$) and MBPP($^+$).}
\label{tab:instruct_humaneval_mbpp}
\end{table}
\subsection{Benchmarks}
\label{sec:benchmarks}
We evaluate the base and instruction models on a diverse suite of coding benchmarks spanning function-level code generation, execution-centric code reasoning, multilingual generalization, and instruction-following code editing:
\begin{itemize}
  \item \textbf{HumanEval / HumanEval+}: HumanEval \citep{chen2021evaluatinglargelanguagemodels} contains 164 Python function-completion tasks evaluated by unit tests. HumanEval+ from EvalPlus \citep{liu2023your} substantially expands the test suites (about $80\times$) for stricter functional correctness. We use EvalPlus to report results on both HumanEval and HumanEval+.

  \item \textbf{MBPP / MBPP+}: MBPP \citep{austin2021program} includes 974 crowd-sourced Python programming problems with tests. We adopt the human-verified EvalPlus subset (399 well-formed tasks) and additionally evaluate on MBPP+, which augments the tests (about $35\times$) for more reliable pass@1 estimation \citep{liu2023your}.

  \item \textbf{CRUXEval}: Small token-level changes can drastically alter program behavior, making code reasoning sensitive and precise. CRUXEval \citep{gu2024cruxeval} provides 800 Python functions with I/O examples and evaluates two tasks: CRUXEval-I (predict inputs given outputs) and CRUXEval-O (predict outputs given inputs).

  \item \textbf{MultiPL-E}: Beyond Python, we evaluated base models on code generation across multiple programming languages using MultiPL-E \citep{cassano2023multiple}, which extends HumanEval to 18 languages. Following Qwen2.5-Coder, we report results on eight representative mainstream languages to assess cross-language coding performance and language-specific effects.

  \item \textbf{MHPP}: To better separate strong models beyond saturated benchmarks (e.g., HumanEval/MBPP), we include MHPP \citep{dai2024mhppexploringcapabilitieslimitations}, which targets harder Python problems with more complex specifications and reasoning requirements.

  \item \textbf{BigCodeBench}: BigCodeBench \citep{zhuo2025bigcodebench} focuses on challenging, realistic programming by requiring tool-like function calls from 139 libraries across 7 domains, covering 1,140 Python tasks with rich context. Each task has multiple tests (5.6 on average) with high branch coverage, and we report results on both the full set and the Hard split of BigCodeBench-Completion.

  \item \textbf{LiveCodeBench}: To mitigate contamination and overfitting to static benchmarks, LiveCodeBench \citep{livecodebench} continuously collects time-stamped problems from competitive programming platforms (e.g., LeetCode, AtCoder, Codeforces), enabling evaluation within recent, user-specified time windows. We follow the v5 evaluation benchmark used by Seed-Coder for comparison.

  \item \textbf{MBXP}: MBXP \citep{athiwaratkun2023multilingual} translates MBPP problems and unit tests into 10+ programming languages, enabling execution-based multilingual evaluation. We report results across 13 widely used languages.

  \item \textbf{NaturalCodeBench}: NaturalCodeBench \citep{athiwaratkun2022multi} contains 402 high-quality problems curated from real user queries (Python and Java) across six practical domains (e.g., software engineering, data science, system administration), with more diverse and complex inputs than classic algorithmic benchmarks.

  \item \textbf{Aider}: We use Aider's \footnote{https://aider.chat/docs/leaderboards/edit.html} code editing benchmark to assess instruction-following edits on existing codebases, based on 133 Exercism exercises, requiring models to produce changes that can be applied automatically and pass tests.

  \item \textbf{CanItEdit}: CanItEdit \citep{cassano2024can} evaluates instructional code editing with 105 hand-crafted problems covering both detailed and underspecified (“lazy”) instructions, testing robustness across diverse edit scenarios.
\end{itemize}

\subsection{Evaluation of Base Models}

\subsubsection{Code Generation}
\paragraph{\textbf{Humaneval and MBPP}} As shown in Table~\ref{tab:base_humaneval_mbpp}, we evaluate \ourmodel-8B-Base using the EvalPlus chat-style prompting template. Among diffusion language models (DLLMs) of comparable scale (approximately 8B parameters), our base model achieves the best overall performance on both HumanEval($^+$) and MBPP($^+$). When compared with similarly sized autoregressive (AR) baselines, \ourmodel consistently outperforms them on HumanEval and HumanEval+, as well as on MBPP, while being only slightly inferior on MBPP+. Notably, \ourmodel exhibits a substantial improvement over our AR baseline, Seed-Coder-8B-Base, across all evaluated HumanEval and MBPP settings.

\paragraph{\textbf{MBXP}} Beyond Python-centric benchmarks, we further evaluate multilingual code generation on MultiPL-E (Table~\ref{tab:base_multipl-e}). \ourmodel{} yields particularly large gains in languages such as C\# and PHP. Since these languages are sparsely represented in our training corpus, we hypothesize that diffusion-style stochastic sampling can effectively amplify learning signals from low-resource code by exposing the model to multiple corrupted-and-denoised views of the same underlying example, thereby improving generalization in data-scarce languages.

\subsubsection{Code Reasoning}
\paragraph{\textbf{CRUXEval}} We additionally assess code reasoning on CRUXEval (Table~\ref{tab:base_bigcodebench_cruxeval}), which consists of Python-only problems. We observe that \ourmodel outperforms Seed-Coder-Base on both reasoning inputs (Input-CoT) and reasoning outputs (Output-CoT). This suggests that incorporating a moderate degree of random masking objectives can effectively enhance the model’s reasoning capability. Moreover, the inputs and outputs in CRUXEval are inherently structured rather than strictly following left-to-right causal logic. As a result, DLLMs benefit from any-order modeling, which enables them to more comprehensively capture the relationships among these structured components.

\subsection{Evaluation of Instruction Models}

\begin{table}[ht]
    \centering
    \resizebox{0.75\textwidth}{!}{
    \begin{tabular}{lr|cc|c|c}
    \toprule
    \multirow{2}{*}{\textbf{Model}} & \multicolumn{1}{r|}{\multirow{2}{*}{\textbf{Size}}} & \multicolumn{1}{c|}{\textbf{MHPP}} & \multicolumn{2}{c|}{\textbf{BigCodeBench}} & \textbf{LiveCodeBench} \\
     & \multicolumn{1}{c|}{} & \multicolumn{1}{c|}{\textit{pass@1}} & \multicolumn{1}{c}{\textit{\ \ \ Full}} & \multicolumn{1}{c|}{\textit{Hard}} & \textit{pass@1} \\ \midrule
    \multicolumn{6}{c}{\textasciitilde 8B Models} \\ \midrule
    CodeLlama-7B-Instruct & \multicolumn{1}{r|}{7B}  & \multicolumn{1}{c|}{6.7} & \multicolumn{1}{c}{\ \ \ 25.7} & \multicolumn{1}{c|}{4.1} & \multicolumn{1}{c}{3.6} \\
    DeepSeek-Coder-6.7B-Instruct & \multicolumn{1}{r|}{6.7B}  & \multicolumn{1}{c|}{20.0} & \multicolumn{1}{c}{\ \ \ 43.8} & \multicolumn{1}{c|}{15.5} & \multicolumn{1}{c}{9.6} \\
    CodeQwen1.5-7B-Chat & \multicolumn{1}{r|}{7B}  & \multicolumn{1}{c|}{17.6} & \multicolumn{1}{c}{\ \ \ 43.6} & \multicolumn{1}{c|}{15.5} & \multicolumn{1}{c}{3.0} \\
    Yi-Coder-9B-Chat & \multicolumn{1}{r|}{9B}  & \multicolumn{1}{c|}{26.7} & \multicolumn{1}{c}{\ \ \ 49.0} & \multicolumn{1}{c|}{17.6} & \multicolumn{1}{c}{17.5} \\
    Llama-3.1-8B-Instruct & \multicolumn{1}{r|}{8B}  & \multicolumn{1}{c|}{17.1} & \multicolumn{1}{c}{\ \ \ 40.5} & \multicolumn{1}{c|}{13.5} & \multicolumn{1}{c}{11.5} \\
    OpenCoder-8B-Instruct & \multicolumn{1}{r|}{8B}  & \multicolumn{1}{c|}{30.5} & \multicolumn{1}{c}{\ \ \ 50.9} & \multicolumn{1}{c|}{18.9} & \multicolumn{1}{c}{17.1} \\
    Qwen2.5-Coder-7B-Instruct & \multicolumn{1}{r|}{7B}  & \multicolumn{1}{c|}{26.7} & \multicolumn{1}{c}{\ \ \ 48.8} & \multicolumn{1}{c|}{20.3} & \multicolumn{1}{c}{17.3} \\
    Qwen3-8B & \multicolumn{1}{r|}{8B}  & \multicolumn{1}{c|}{32.8} & \multicolumn{1}{c}{\ \ \ 51.7} & \multicolumn{1}{c|}{23.0} & \multicolumn{1}{c}{23.5} \\
    Seed-Coder-8B-Instruct & \multicolumn{1}{r|}{8B}  & \multicolumn{1}{c|}{36.2} & \multicolumn{1}{c}{\ \ \ 53.3} & \multicolumn{1}{c|}{26.4} & \multicolumn{1}{c}{\fst{24.7}} \\  
    \graybg \ourmodel-8B-Instruct & \multicolumn{1}{r|}{8B}  & \multicolumn{1}{c|}{\fst{42.4}} & \multicolumn{1}{c}{\ \ \ \fst{54.8}} & \multicolumn{1}{c|}{\fst{31.8}} & \multicolumn{1}{c}{23.5} \\ 
    \midrule
    \multicolumn{6}{c}{\grayt 13B+ Models and External API} \\ \midrule
    \grayt StarCoder2-15B-Instruct & \multicolumn{1}{r|}{\grayt 15B} & \multicolumn{1}{c|}{\grayt 19.0} & \multicolumn{1}{c}{\ \ \ \grayt 45.1} & \multicolumn{1}{c|}{\grayt 14.9} & \multicolumn{1}{c}{\grayt 5.3} \\
    \grayt Codestral-22B & \multicolumn{1}{r|}{\grayt 22B}  & \multicolumn{1}{c|}{\grayt 25.2} & \multicolumn{1}{c}{\ \ \ \grayt 52.5} & \multicolumn{1}{c|}{\grayt 24.3} & \multicolumn{1}{c}{\grayt 20.5} \\
    \grayt CodeLlama-70B-Instruct & \multicolumn{1}{r|}{\grayt 70B}  & \multicolumn{1}{c|}{\grayt 19.5} & \multicolumn{1}{c}{\ \ \ \grayt 49.6} & \multicolumn{1}{c|}{\grayt 15.5} & \multicolumn{1}{c}{\grayt 14.5} \\
    \grayt DeepSeek-Coder-33B-Instruct & \multicolumn{1}{r|}{\grayt 33B}  & \multicolumn{1}{c|}{\grayt 32.9} & \multicolumn{1}{c}{\ \ \ \grayt 51.1} & \multicolumn{1}{c|}{\grayt 20.9} & \multicolumn{1}{c}{\grayt 14.5} \\ 
    \grayt DeepSeek-Coder-V2-Lite-Instruct & \multicolumn{1}{r|}{\grayt 2.4B/16B}  & \multicolumn{1}{c|}{\grayt 30.5} & \multicolumn{1}{c}{\ \ \ \grayt 47.6} & \multicolumn{1}{c|}{\grayt 18.2} & \multicolumn{1}{c}{\grayt 14.2} \\
    \grayt DeepSeek-Coder-V2-Instruct & \multicolumn{1}{r|}{\grayt 21B/236B}  & \multicolumn{1}{c|}{\grayt 31.9} & \multicolumn{1}{c}{\ \ \ \grayt 59.7} & \multicolumn{1}{c|}{\grayt 33.1} & \multicolumn{1}{c}{\grayt 28.9} \\
    \grayt Qwen2.5-Coder-14B-Instruct & \multicolumn{1}{r|}{\grayt 14B} & \multicolumn{1}{c|}{\grayt 36.7} & \multicolumn{1}{c}{\ \ \ \grayt 52.2} & \multicolumn{1}{c|}{\grayt 16.2} & \multicolumn{1}{c}{\grayt 19.3} \\
    \grayt Qwen2.5-Coder-32B-Instruct & \multicolumn{1}{r|}{\grayt 32B}  & \multicolumn{1}{c|}{\grayt \fst{42.4}} & \multicolumn{1}{c}{\ \ \ \grayt \fst{52.3}} & \multicolumn{1}{c|}{\grayt \fst{20.9}} & \multicolumn{1}{c}{\grayt \fst{30.7}} \\
    LLaDA2.0-flash & \multicolumn{1}{r|}{6B/100B}  & \multicolumn{1}{c|}{-} & \multicolumn{1}{c}{\ \ \ 41.6} & \multicolumn{1}{c|}{-} & \multicolumn{1}{c}{-} \\
    Seed-Diffusion-Preview(0705) & \multicolumn{1}{r|}{-}  & \multicolumn{1}{c|}{-} & \multicolumn{1}{c}{\ \ \ 53.2} & \multicolumn{1}{c|}{-} & \multicolumn{1}{c}{-} \\
    Mercury Coder & \multicolumn{1}{r|}{-}  & \multicolumn{1}{c|}{-} & \multicolumn{1}{c}{\ \ \ 45.5} & \multicolumn{1}{c|}{-} & \multicolumn{1}{c}{-} \\
    Gemini Diffusion & \multicolumn{1}{r|}{-}  & \multicolumn{1}{c|}{-} & \multicolumn{1}{c}{\ \ \ 45.4} & \multicolumn{1}{c|}{-} & \multicolumn{1}{c}{-} \\
    \bottomrule
    \end{tabular}
    }
        \captionsetup{justification=justified, singlelinecheck=false}
    \caption{Performance of instruct models on MHPP, BigCodeBench-Completion and LiveCodeBench (v5).}
    \label{tab:instruct-codegen}
    \end{table}
\begin{table}[!h]
\centering
\resizebox{\textwidth}{!}{
\begin{tabular}{lrccccccccccccc|c}
\toprule
\textbf{Model} & \multicolumn{1}{r|}{\textbf{Size}} & Python & Java & C++ & C\# & TS & JS & PHP & Go & Kotlin & Perl & Ruby & Scala & Swift & \textbf{Average} \\ \midrule
\multicolumn{16}{c}{\textasciitilde 8B Models} \\ \midrule
CodeLlama-7B-Instruct & \multicolumn{1}{r|}{7B} & 54.0 & 38.8 & 32.9 & 50.0 & 42.3 & 45.5 & 36.6 & 48.8 & 47.2 & 50.1 & 36.9 & 40.2 & 33.2 & 42.8 \\
DeepSeek-Coder-6.7B-Instruct & \multicolumn{1}{r|}{6.7B} & 74.9 & 52.2 & 30.9 & 55.9 & 64.8 & 64.7 & 25.8 & 93.8 & 59.6 & 3.3 & 65.9 & 54.8 & 47.4 & 53.4 \\
CodeQwen1.5-7B-Chat & \multicolumn{1}{r|}{7B} & 77.7 & 66.6 & 66.8 & 64.4 & 66.7 & 67.5 & 67.3 & 55.1 & 60.9 & 61.1 & 65.9 & 60.0 & 54.7 & 64.2 \\
Yi-Coder-9B-Chat & \multicolumn{1}{r|}{9B} & 82.0 & 73.4 & 79.1 & 70.3 & 74.1 & 73.3 & 76.4 & 90.9 & 64.4 & 60.9 & 67.3 & 63.5 & 57.3 & 71.8 \\
Llama-3.1-8B-Instruct & \multicolumn{1}{r|}{8B} & 70.1 & 59.8 & 59.1 & 56.6 & 59.1 & 59.1 & 62.5 & 85.7 & 52.2 & 42.6 & 55.9 & 44.5 & 31.8 & 56.8 \\
OpenCoder-8B-Instruct & \multicolumn{1}{r|}{8B} & 79.1 & 68.1 & 71.3 & 71.0 & 67.6 & 61.4 & 68.1 & 94.4 & 66.4 & 56.1 & 70.5 & 63.1 & 56.7 & 68.8 \\
Qwen2.5-Coder-7B-Instruct & \multicolumn{1}{r|}{7B} & 83.5 & 70.5 & 74.1 & 71.5 & 72.2 & 74.1 & 74.2 & 96.0 & 65.5 & 64.4 & 75.5 & 64.2 & \fst{62.0} & 72.9 \\
Qwen3-8B & \multicolumn{1}{r|}{8B} & 77.0 & 69.0 & 72.8 & 68.9 & 73.0 & 73.8 & 72.3 & 92.9 & 62.0 & 64.6 & 69.0 & 63.1 & 42.2 & 69.3 \\
Seed-Coder-8B-Instruct & \multicolumn{1}{r|}{8B} & 85.2 & 72.7 & 77.0 & \fst{74.2} & 72.8 & \fst{78.8} & \fst{74.7} & 95.5 & \fst{73.4} & \fst{72.5} & \fst{78.0} & 70.3 & 54.2 & \fst{75.3} \\
\graybg \ourmodel-8B-instruct & \multicolumn{1}{r|}{8B} & \fst{85.7} & \fst{75.3} & \fst{77.8} & 71.2 & \fst{73.0} & 76.9 & 73.8 & \fst{98.7} & 72.5 & 70.5 & 77.1 & \fst{71.4} & 54.2 & \fst{75.3}\\
\midrule
\multicolumn{16}{c}{\grayt 13B+ Models and External API} \\ \midrule
\grayt StarCoder2-15B-Instruct & \multicolumn{1}{r|}{\grayt 15B} & \grayt 78.0 & \grayt 25.1 & \grayt 25.9 & \grayt 21.7 & \grayt 20.7 & \grayt 59.8 & \grayt 53.5 & \grayt 90.4 & \grayt 46.7 & \grayt 31.9 & \grayt 56.1 & \grayt 43.2 & \grayt 42.0 & \grayt 45.8 \\
\grayt Codestral-22B & \multicolumn{1}{r|}{\grayt 22B} & \grayt 78.2 & \grayt 73.6 & \grayt 77.3 & \grayt 70.1 & \grayt 71.7 & \grayt 68.5 & \grayt 74.9 & \grayt 97.1 & \grayt 71.0 & \grayt 66.6 & \grayt 74.2 & \grayt 64.4 & \grayt 50.1 & \grayt 72.1 \\
\grayt CodeLlama-70B-Instruct & \multicolumn{1}{r|}{\grayt 70B} & \grayt 77.8 & \grayt 66.6 & \grayt 68.6 & \grayt 69.2 & \grayt 47.8 & \grayt 62.5 & \grayt 70.5 & \grayt 77.7 & \grayt 57.2 & \grayt 51.1 & \grayt 67.0 & \grayt 51.3 & \grayt 48.7 & \grayt 62.8 \\
\grayt DeepSeek-Coder-33B-Instruct & \multicolumn{1}{r|}{\grayt 33B} & \grayt 80.4 & \grayt 71.8 & \grayt 76.8 & \grayt 69.9 & \grayt 72.4 & \grayt 69.8 & \grayt 75.1 & \grayt 96.4 & \grayt 70.1 & \grayt 66.6 & \grayt 75.1 & \grayt 64.6 & \grayt 54.3 & \grayt 72.6 \\
\grayt DeepSeek-Coder-V2-Lite-Instruct & \multicolumn{1}{r|}{\grayt 2.4B/16B} & \grayt 82.8 & \grayt 73.3 & \grayt 75.3 & \grayt 72.4 & \grayt 72.4 & \grayt 73.1 & \grayt 75.1 & \grayt 95.1 & \grayt 69.9 & \grayt 61.6 & \grayt 74.5 & \grayt 63.5 & \grayt 55.0 & \grayt 72.6 \\
\grayt DeepSeek-Coder-V2-Instruct & \multicolumn{1}{r|}{\grayt 21B/236B} & \grayt 89.4 & \grayt 78.2 & \grayt 77.6 & \grayt 72.6 & \grayt 74.8 & \grayt \fst{80.5} & \grayt 75.8 & \grayt 89.1 & \grayt 74.5 & \grayt 70.7 & \grayt 80.2 & \grayt 67.9 & \grayt 59.0 & \grayt 76.2 \\
\grayt Qwen2.5-Coder-14B-Instruct & \multicolumn{1}{r|}{\grayt 14B} & \grayt 86.2 & \grayt 77.5 & \grayt 84.8 & \grayt \fst{80.1} & \grayt 77.6 & \grayt 77.7 & \grayt 79.7 & \grayt \fst{97.1} & \grayt 75.3 & \grayt \fst{76.2} & \grayt 79.3 & \grayt \fst{73.1} & \grayt \fst{67.2} & \grayt 79.4 \\
\grayt Qwen2.5-Coder-32B-Instruct & \multicolumn{1}{r|}{\grayt 32B} & \grayt \fst{90.2} & \grayt \fst{80.4} & \grayt \fst{86.3} & \grayt 73.5 & \grayt \fst{78.3} & \grayt 79.3 & \grayt \fst{87.6} & \grayt 96.4 & \grayt \fst{75.6} & \grayt 74.7 & \grayt \fst{83.4} & \grayt 63.3 & \grayt 66.7 & \grayt \fst{79.7} \\
Seed-Diffusion-Preview(0705) & \multicolumn{1}{r|}{-} & 79.4 & 67.7 & 72.6 & 70.3 & 73.0 & 76.6 & 74.7 & 92.9 & 71.2 & 71.2 & 72.5 & 67.0 & 54.2 & 72.6 \\
\bottomrule
\end{tabular}
}
\caption{Performance of instruct models on MBXP.}
\label{tab:instruct_mbxp}
\end{table}
\begin{table}[!h]
\centering
\resizebox{0.85\textwidth}{!}{
\begin{tabular}{lr|cccccc|c}
\toprule
\multirow{2}{*}{\textbf{Model}} & \multicolumn{1}{r|}{\multirow{2}{*}{\textbf{Size}}} & \multicolumn{3}{c}{NCB (zh)} & \multicolumn{3}{c|}{NCB (en)} & \multirow{2}{*}{\textbf{Total}} \\ \cmidrule(lr){3-5} \cmidrule(lr){6-8}
 & \multicolumn{1}{r|}{} & \textit{Python} & \textit{Java} & \multicolumn{1}{l}{\textit{Total}} & \textit{Python} & \textit{Java} & \multicolumn{1}{c}{\textit{Total}} & \multicolumn{1}{|c}{} \\ \midrule
\multicolumn{9}{c}{\textasciitilde 8B Models} \\ \midrule
CodeLlama-7B-Instruct & 7B & 18.6 & 8.6 & 13.6 & 17.1 & 14.3 & 15.7 & 14.6 \\
DeepSeek-Coder-6.7B-Instruct & 6.7B & 38.6 & 31.4 & 35.0 & 32.9 & 32.9 & 32.9 & 33.9 \\
CodeQwen1.5-7B-Chat & 7B & 30.0 & 28.6 & 29.3 & 30.0 & 27.1 & 28.6 & 25.7 \\
Yi-Coder-9B-Chat & 9B & 41.4 & \fst{45.7} & 43.6 & 38.6 & 44.3 & 41.5 & 42.5 \\
Llama-3.1-8B-Instruct & 8B & 27.1 & 24.3 & 25.7 & 22.9 & 22.9 & 22.9 & 24.3 \\
OpenCoder-8B-Instruct & 8B & 40.0 & 30.0 & 35.0 & 35.7 & 24.3 & 30.0 & 32.5 \\
Qwen2.5-Coder-7B-Instruct & 7B & 34.3 & 37.1 & 35.7 & 34.3 & 35.7 & 35.0 & 35.4 \\
Qwen3-8B & 8B & 37.1 & 32.9 & 35.0 & 34.3 & 38.6 & 36.5 & 35.7 \\
Seed-Coder-8B-Instruct & 8B & \fst{55.7} & \fst{45.7} & \fst{50.7} & \fst{50.0} & \fst{47.1} & \fst{48.6} & \fst{49.6} \\ 
\graybg \ourmodel-8B-Instruct & 8B & 51.4 & \fst{45.7} & 48.6 & \fst{50.0} & \fst{47.1} & \fst{48.6} & 48.6 \\ 
\midrule
\multicolumn{9}{c}{\grayt 13B+ Models and External API} \\ \midrule
\grayt StarCoder2-15B-Instruct & \grayt 15B & \grayt 44.3 & \grayt 30.0 & \grayt 37.2 & \grayt 38.6 & \grayt 42.9 & \grayt 40.8 & \grayt 39.0 \\
\grayt Codestral-22B & \grayt 22B & \grayt 40.0 & \grayt 44.3 & \grayt 42.2 & \grayt 41.4 & \grayt \fst{45.7} & \grayt 43.6 & \grayt 42.9 \\
\grayt CodeLlama-70B-Instruct & \grayt 70B & \grayt 35.1 & \grayt 32.1 & \grayt 33.6 & \grayt 32.8 & \grayt 30.5 & \grayt 31.7 & \grayt 32.6 \\
\grayt DeepSeek-Coder-33B-Instruct & \grayt 33B & \grayt 44.3 & \grayt 38.9 & \grayt 41.6 & \grayt \fst{44.3} & \grayt 44.3 & \grayt \fst{44.3} & \grayt 43.0 \\
\grayt DeepSeek-Coder-V2-Lite-Instruct & \grayt 2.4B/16B & \grayt 41.4 & \grayt 47.1 & \grayt 44.3 & \grayt 41.4 & \grayt 37.1 & \grayt 39.3 & \grayt 41.8 \\
\grayt Qwen2.5-Coder-14B-Instruct & \grayt 14B & \grayt \fst{48.6} & \grayt \fst{48.6} & \grayt \fst{48.6} & \grayt 42.9 & \grayt \fst{45.7} & \grayt \fst{44.3} & \grayt \fst{46.4} \\
Seed-Diffusion-Preview(0705) & - & 52.9 & 38.6 & 45.8 & 45.7 & 38.6 & 42.2 & 43.9 \\ 
\bottomrule
\end{tabular}
}
\caption{Performance of instruct models on NaturalCodeBench.}
\label{tab:instruct_naturalcodebench}
\end{table}

\begin{table}[!h]
\centering
\resizebox{0.6\textwidth}{!}{
\begin{tabular}{lrcc}
\toprule
\textbf{Model} & \multicolumn{1}{r|}{\textbf{Size}} & Input-CoT & Output-CoT \\ \midrule
\multicolumn{4}{c}{\textasciitilde 8B Models} \\ \midrule
CodeLlama-7B-Instruct & \multicolumn{1}{r|}{7B} & 36.1 & 36.2 \\
DeepSeek-Coder-6.7B-Instruct & \multicolumn{1}{r|}{6.7B} & 42.6 & 45.1 \\
CodeQwen1.5-7B-Chat & \multicolumn{1}{r|}{7B} & 44.0 & 38.8 \\
Yi-Coder-9B-Chat & \multicolumn{1}{r|}{9B} & 47.5 & 55.6 \\
Llama-3.1-8B-Instruct & \multicolumn{1}{r|}{8B} & 35.6 & 37.8 \\
OpenCoder-8B-Instruct & \multicolumn{1}{r|}{8B} & 39.9 & 43.0 \\
Qwen2.5-Coder-7B-Instruct  & \multicolumn{1}{r|}{7B} & 65.8 & 65.9 \\
Qwen3-8B & \multicolumn{1}{r|}{8B} & \fst{73.8} & \fst{76.9} \\
Seed-Coder-8B-Instruct & \multicolumn{1}{r|}{8B} & 63.3 & 67.1 \\ 
\graybg \ourmodel-8B-Instruct & \multicolumn{1}{r|}{8B} & 62.1 & 69.0 \\ 
\midrule
\multicolumn{4}{c}{\grayt 13B+ Models} \\ \midrule
\grayt StarCoder2-15B-Instruct & \multicolumn{1}{r|}{\grayt 15B} & \grayt 45.5 & \grayt 60.9 \\
\grayt Codestral-22B & \multicolumn{1}{r|}{\grayt 22B} & \grayt 61.3 & \grayt 63.5 \\
\grayt CodeLlama-70B-Instruct & \multicolumn{1}{r|}{\grayt 70B} & \grayt 56.5 & \grayt 57.8 \\
\grayt DeepSeek-Coder-33B-Instruct & \multicolumn{1}{r|}{\grayt 33B} & \grayt 47.3 & \grayt 50.6 \\
\grayt DeepSeek-Coder-V2-Lite-Instruct & \multicolumn{1}{r|}{\grayt 2.4B/16B} & \grayt 53.0 & \grayt 52.9 \\
\grayt Qwen2.5-Coder-14B-Instruct & \multicolumn{1}{r|}{\grayt 14B} & \grayt \fst{69.5} & \grayt \fst{79.5} \\
\bottomrule
\end{tabular}
}
\caption{Performance of instruct models on CRUXEval.}
\label{tab:instruct_cruxeval}
\end{table}
\begin{table}[!h]
\centering
\resizebox{0.65\textwidth}{!}{
\begin{tabular}{lrcc}
\toprule
\multirow{2}{*}{\textbf{Model}} & \multicolumn{1}{r|}{\multirow{2}{*}{\textbf{Size}}} & \multicolumn{1}{c|}{\textbf{Aider}} & \multicolumn{1}{c}{\textbf{CanItEdit}} \\
 & \multicolumn{1}{c|}{} & \multicolumn{1}{c|}{\textit{tries=2}} & \textit{pass@1} \\ \midrule
\multicolumn{4}{c}{\textasciitilde 8B Models} \\ \midrule
CodeLlama-7B-Instruct & \multicolumn{1}{r|}{7B} & \multicolumn{1}{c|}{1.5} & 25.7 \\
DeepSeek-Coder-6.7B-Instruct & \multicolumn{1}{r|}{6.7B} & \multicolumn{1}{c|}{44.4} & 36.9 \\
CodeQwen1.5-7B-Chat & \multicolumn{1}{r|}{7B} & \multicolumn{1}{c|}{38.3} & 34.8 \\
Yi-Coder-9B-Chat & \multicolumn{1}{r|}{9B} & \multicolumn{1}{c|}{54.1} & 50.5 \\
Llama-3.1-8B-Instruct & \multicolumn{1}{r|}{8B} & \multicolumn{1}{c|}{33.1} & 39.5 \\
OpenCoder-8B-Instruct & \multicolumn{1}{r|}{8B} & \multicolumn{1}{c|}{30.8} & 39.0 \\
Qwen2.5-Coder-7B-Instruct & \multicolumn{1}{r|}{7B} & \multicolumn{1}{c|}{\fst{57.9}} & 49.5 \\
Qwen3-8B & \multicolumn{1}{r|}{8B} & \multicolumn{1}{c|}{55.6} & 45.7 \\
Seed-Coder-8B-Instruct & \multicolumn{1}{r|}{8B} & \multicolumn{1}{c|}{57.1} & 50.5 \\ 
\graybg \ourmodel-8B-Instruct & \multicolumn{1}{r|}{8B} & \multicolumn{1}{c|}{54.9} & \fst{60.0} \\ 
\midrule
\multicolumn{4}{c}{\grayt 13B+ Models and External API} \\ \midrule
\grayt StarCoder2-15B-Instruct & \multicolumn{1}{r|}{\grayt 15B} & \multicolumn{1}{c|}{\grayt 38.2} & \grayt 31.4 \\
\grayt Codestral-22B & \multicolumn{1}{r|}{\grayt 22B} & \multicolumn{1}{c|}{\grayt 51.1} & \grayt 52.4 \\
\grayt CodeLlama-70B-Instruct & \multicolumn{1}{r|}{\grayt 70B} & \multicolumn{1}{c|}{\grayt 15.0} & \grayt 40.5 \\
\grayt DeepSeek-Coder-33B-Instruct & \multicolumn{1}{r|}{\grayt 33B} & \multicolumn{1}{c|}{\grayt 54.5} & \grayt 46.2 \\
\grayt DeepSeek-Coder-V2-Lite-Instruct & \multicolumn{1}{r|}{\grayt 2.4B/16B} & \multicolumn{1}{c|}{\grayt 52.6} & \grayt 45.2 \\
\grayt Qwen2.5-Coder-14B-Instruct & \multicolumn{1}{r|}{\grayt 14B} & \multicolumn{1}{c|}{\grayt \fst{69.2}} & 52.9 \\
Seed-Diffusion-Preview(0705) & \multicolumn{1}{r|}{-} & \multicolumn{1}{c|}{44.4} & \grayt{54.3} \\ 
\bottomrule
\end{tabular}
}
\caption{Performance of instruct models on Aider (``whole'' format) and CanItEdit.}
\label{tab:instruct_aider_canitedit}
\end{table}

\subsubsection{Code Generation}
\paragraph{\textbf{Humaneval and MBPP.}} On standard function-level code generation benchmarks, \ourmodel-8B-Instruct achieves strong gains over the autoregressive baseline.
As shown in Table~\ref{tab:instruct_humaneval_mbpp}, \ourmodel-8B-Instruct significantly improves upon Seed-Coder-8B-Instruct on both HumanEval(+) and MBPP(+), and on MBPP it outperforms all other instruction models while clearly surpassing all $\sim$8B-scale diffusion models.

\paragraph{\textbf{MHPP.}} On the more challenging MHPP benchmark, which only reveals scores through an official submission interface to mitigate data contamination and ensure fair comparison, \ourmodel-8B-Instruct attains the best performance among all compared models and reaches the level of Qwen2.5-Coder-32B-Instruct (Table~\ref{tab:instruct-codegen}).

\paragraph{\textbf{BigCodeBench.}} For more realistic programming tasks, BigCodeBench measures the ability to solve challenging, real-world coding problems with rich context and tool-like function calls.
As shown in Table~\ref{tab:instruct-codegen}, \ourmodel-8B-Instruct delivers substantial improvements over Seed-Coder-8B-Instruct and, among all models, is only surpassed by the much larger DeepSeek-Coder-V2-Instruct (21B/236B), demonstrating the strong practical code generation capabilities of \ourmodel-8B-Instruct.

\paragraph{\textbf{LiveCodeBench.}} To ensure a fair comparison under identical data and training settings, we adopt exactly the same evaluation configuration as Seed-Coder-8B-Instruct on the LiveCodeBench (v5, \text{2024.10-2025.02}).
Here \ourmodel-8B-Instruct (23.5\%) is slightly behind Seed-Coder-8B-Instruct (24.7\%), but matches the performance of Qwen3-8B and remains stronger than other $\sim$8B-scale models (Table~\ref{tab:instruct-codegen}).

\paragraph{\textbf{MBXP.}} We further evaluate multilingual code generation with MBXP.
As summarized in Table~\ref{tab:instruct_mbxp}, under the 10+ programming language setting \ourmodel-8B-Instruct achieves an overall average comparable to Seed-Coder-8B-Instruct, while attaining the highest scores in most languages among $\sim$8B instruct models. Due to the need to extensively supplement the scarce data such as C\# and PHP that are lacking in pretraining during SFT, the advantage in multilingual coding capabilities has been reduced.

\paragraph{\textbf{NaturalCodeBench.}} On NaturalCodeBench, which targets practical software engineering problems in Python and Java, \ourmodel-8B-Instruct is slightly weaker than Seed-Coder-8B-Instruct on Chinese Python queries but is otherwise on par overall (Table~\ref{tab:instruct_naturalcodebench}). Both models, however, are clearly ahead of other $\sim$8B models and even competitive with many 13B+ and external API systems.

\subsubsection{Code Reasoning}

\paragraph{\textbf{CRUXEval.}} For execution-centric code reasoning, we evaluate on CRUXEval.
Consistent with the trends observed for base models, Table~\ref{tab:instruct_cruxeval} shows that \ourmodel-8B-Instruct achieves stronger performance than Seed-Coder-8B-Instruct on the Output-CoT setting and slightly better average performance across Input-CoT and Output-CoT.
Nonetheless, Qwen3-8B remains ahead on CRUXEval, indicating that there is still considerable headroom for specialized small-code models on fine-grained reasoning tasks.

\subsubsection{Code editing}

\paragraph{\textbf{CanItEdit.}} On CanItEdit, \ourmodel-8B-Instruct substantially outperforms all other models (Table~\ref{tab:instruct_aider_canitedit}).
We hypothesize that this gain benefits from the denoising nature of DLLMs: random masking and reconstruction inherently train the model on edit- and infill-like patterns, enabling it to better exploit editing supervision and extract more editing-related knowledge from the same data.

\paragraph{\textbf{Aider.}} On Aider, which evaluates multi-turn editing and reasoning over entire codebases, \ourmodel-8B-Instruct remains slightly weaker than Seed-Coder-8B-Instruct under the tries=2 setting (Table~\ref{tab:instruct_aider_canitedit}).
This task requires concatenating outputs across turns, often yielding very long contexts that exceed the 8192-token window used during training, and we observe a mild performance drop in this regime.
Even so, \ourmodel-8B-Instruct reaches performance comparable to Qwen3-8B and surpasses larger models such as DeepSeek-Coder-33B-Instruct, indicating that it still offers strong practical code editing abilities despite this limitation.

\section{Conclusion, Limitation, and Future Work}
In this report, we present \ourmodel, which serves as a practical best-practice attempt demonstrating that the training paradigm of diffusion language models can provide effective data augmentation and lead to improved model performance.
By analyzing how to efficiently enhance knowledge learning in DLLMs, designing a well-aligned training pipeline, and incorporating techniques that stabilize and optimize the learning process, we show that—while keeping the architecture and data identical to Seed-Coder—the diffusion-based training procedure yields consistently better results.
These findings further indicate that the sampling process of text diffusion models can function as a principled and effective form of data augmentation for model training.
On comprehensive code evaluation benchmarks, \ourmodel achieves state-of-the-art results among$\sim$8B AR-based or diffusion-based code models.

Since \ourmodel is primarily focused on the code domain and lacks large-scale training data from other areas, its performance on mathematical reasoning and general-purpose text tasks may be relatively limited.
Whether text diffusion sampling can provide even greater benefits in broader domains remains an open question, requiring future model iterations and deeper empirical exploration.

\clearpage
\newpage

\newcommand{\contribrole}[2]{%
  \noindent\textbf{#1}\\
  #2\par\vspace{0.6em}
}
\section{Contributions} \label{sec:contributions}
\setlength{\parindent}{0em}

\contribrole{Project Lead}{%
  Chenghao Fan\textsuperscript{1,2}
}

\contribrole{Core Contributor}{%
  Chenghao Fan\textsuperscript{1,2}
}

\contribrole{Contributor$^*$}{%
  Wen Heng\textsuperscript{2},
  Bo Li\textsuperscript{2},
  Sichen Liu\textsuperscript{1},
  Yuxuan Song\textsuperscript{2}
  Jing Su\textsuperscript{2},
  
}

\contribrole{Supervision$^*$}{%
  Wen Heng\textsuperscript{2},
  Xiaoye Qu\textsuperscript{1},
 Kai Shen\textsuperscript{2},
 Wei Wei\textsuperscript{1},
}

($^*$Sorted Alphabetically)

\contribrole{Affiliation}{%
\textsuperscript{1}School of Computer Science \& Technology, Huazhong University of Science and Technology\\
\textsuperscript{2}ByteDance Seed
}

\section{Acknowledgments}
We gratefully acknowledge and thank every Seed-LLM team member not explicitly mentioned above. We also acknowledge and thank every Seed team member for the valuable support. We thank Shulin Xin, Qi Liu, Yirong Chen, Zhexi Zhang, Ziwen Xu, Shen Nie, Hongrui Zhan, Shen Zheng for insightful technical discussions.
\clearpage

\bibliographystyle{plainnat}
\bibliography{main}

\clearpage



\end{document}